\documentclass{article}

\usepackage[final]{neurips_2021}

\usepackage[utf8]{inputenc} % allow utf-8 input
\usepackage[T1]{fontenc}    % use 8-bit T1 fontshttps://www.overleaf.com/project/6099b159c3df141cfc0ecdb6
\usepackage{url}            % simple URL typesetting
\usepackage{booktabs}       % professional-quality tables
\usepackage{amsfonts}       % blackboard math symbols
\usepackage{nicefrac}       % compact symbols for 1/2, etc.
\usepackage{microtype}      % microtypography
\usepackage{xcolor}         % colors
\usepackage[numbers]{natbib}

\usepackage[super]{nth}
\usepackage{amsmath,amsfonts,bm}

\def\eqref#1{equation~\ref{#1}}
\def\1{\bm{1}}

\DeclareMathAlphabet{\mathsfit}{\encodingdefault}{\sfdefault}{m}{sl}
\SetMathAlphabet{\mathsfit}{bold}{\encodingdefault}{\sfdefault}{bx}{n}

\newcommand{\E}{\mathbb{E}}

\usepackage{amsmath,amsfonts,bm}
\newcommand{\inputpattern}[0]{\pi_{i}}
\newcommand{\allpatterns}[0]{\Pi}
\newcommand{\positivepatterns}[0]{\Pi_{+}}
\newcommand{\negativepatterns}[0]{\Pi_{-}}
\newcommand{\pmpatterns}[0]{\Pi_{\pm}}

\newcommand{\randompatterns}[0]{\Pi_{\text{rand}}}
\newcommand{\baselineattentionpatterns}[0]{\Pi_{\text{attn}}}
\newcommand{\baselineinfluencepatterns}[0]{\Pi_{\text{inf}}}
\newcommand{\baselineconductancepatterns}[0]{\Pi_{\text{cond}}}

\newcommand{\xvec}{\mathbf{x}}
\newcommand{\hvec}{\mathbf{h}}
\newcommand{\yvec}{\mathbf{y}}

\newcommand{\suchthat}[0]{:}

\newcommand{\fref}[1]{Fig.~\ref{#1}}
\newcommand{\sref}[1]{Sec.~\ref{#1}}

\usepackage{wasysym}
\newtheorem{definition}{Definition}

\newtheorem{proposition}{Proposition}

\newtheorem{remark}{Remark}
\newcommand{\stacklabel}[1]{\stackrel{\smash{\scriptscriptstyle \mathrm{#1}}}}
\newcommand{\defeq}{\stacklabel{def}=}

\newcommand{\sparen}[1]{\left[ #1 \right]}

\newcommand{\set}[1]{\left\{ #1 \right\}}

\newcommand{\rar}{\rightarrow}

\usepackage{babel}
\def\final{} % uncomment this to hide all comments

\ifdefined\final
\else
\def\enablecomments
\fi

\ifdefined\draft
\def\enablecomments{}
\fi

\usepackage{soul}
\usepackage{xcolor}

\definecolor{LightGreen}{rgb}{0.80,1.00,0.80}
\definecolor{LightBlue}{rgb}{0.80,0.80,1.00}
\definecolor{LightRed}{rgb}{1.00,0.80,0.80}
\definecolor{LightPurple}{rgb}{0.94,0.85,1.00}
\definecolor{LightGray}{rgb}{0.90,0.90,0.90}

\soulregister{\method}{7}
\soulregister{\xspace}{7}
\soulregister{\emph}{7}
\soulregister{\ref}{7}

\ifdefined\enablecomments
  \DeclareRobustCommand{\commentformat}[3]{\sethlcolor{#2}\textsf{\hl{#1: #3}}}
  \newcommand{\sm}    [1]{{\scriptsize\sethlcolor{LightGray}\hl{\textsf{#1}}}}
   
\else
  \newcommand{\commentformat}[3]{}
  \newcommand{\sm}    [1]{}
   
\fi

\newcommand{\anupam}[1]{\commentformat{AD}{LightRed}{#1}}

\newcommand{\todo}[1]{\commentformat{TODO}{LightRed}{#1}}

\usepackage{graphicx}
\usepackage{algorithmic}
\usepackage[ruled,vlined]{algorithm2e}

\usepackage{amssymb}
\usepackage{makecell}
\usepackage{multirow}
\usepackage{multicol}
\usepackage{color}
\usepackage{tikz}
\usepackage{subcaption}
\usepackage{printlen}\uselengthunit{in}
\usepackage{bbm}

\usepackage{amsfonts}
\usepackage{booktabs}
\usepackage{siunitx}
\usepackage{adjustbox}
\usepackage{array}

\usepackage{tikz}

\usetikzlibrary{math,snakes,shapes,decorations,arrows,calc,arrows.meta,fit,positioning}
\usepackage{hyperref}       % hyperlinks

\title{Influence Patterns for Explaining Information Flow in BERT}

\author{Kaiji Lu\thanks{Correspondence to \texttt{kaijil@andrew.cmu.edu}},  Zifan Wang,  Piotr Mardziel,  Anupam Datta \\
Electrical and Computer Engineering\\
Carnegie Mellon University\\
Mountain View, CA 94089 \\
}

\begin{document}
\maketitle

\begin{abstract}
While \emph{``attention is all you need''}  may be proving true, we do not know \emph{why}: attention-based transformer models such as BERT are superior but how information flows from input tokens to output predictions are unclear.  We introduce \emph{influence patterns},  abstractions  of  sets  of  paths  through  a  transformer model.  Patterns  quantify  and localize the flow of  information to paths passing through a sequence of model nodes. Experimentally, we find that significant portion of information flow in BERT goes through skip connections instead of attention heads. We further show that consistency of patterns across instances is an indicator of BERT’s performance. Finally, we demonstrate that patterns account for far more model performance than previous attention-based and layer-based methods.
\end{abstract}
\section{Introduction}

% In this paper, we introduce a technique for explaining how transformer models capture linguistic concepts across layers.

Previous works show that transformer models such as BERT~\cite{devlin2018bert} encode various linguistic concepts~\citep{lin2019open,tenney2019bert,goldberg2019assessing}, some of which can be associated with internal components of each layer, such as internal embeddings or attention weights~\citep{lin2019open,tenney2019bert,hewitt2019structural,coenen2019visualizing}. 
However, exactly how information flows through a transformer from input tokens to the output predictions remains an open question. Recent attempts to answer this question include using attention-based methods, where attention weights are used as indicators of flow of information~\citep{clark2019does, abnar2020quantifying, wu2020structured}, or layer-based approaches\cite{hewitt2019structural, hewitt2019designing,coenen2019visualizing}, which identify important network units in each layer. 

In this paper, we examine the information flow question through an alternative lens of gradient-based attribution methods. We introduce \emph{influence patterns}, abstractions of sets of gradient-based paths through a transformer's entire computational graph. We also introduce a greedy search procedure for efficiently and effectively finding patterns representative of concept-critical information flow. Figure~\ref{fig:berts} provides an example of an influence pattern in BERT. We conduct an extensive empirical study of influence patterns for several NLP tasks: Subject-Verb Agreement (SVA), Reflexive Anaphora (RA), and Sentiment Analysis (SA). Our findings are summarized below. 

\anupam{add quantitative numbers + paper sections below}
\begin{itemize}
    \item A significant portion of information flows in BERT go through skip connections and not attention heads, indicating that attention weights\cite{abnar2020quantifying} alone are not sufficient to characterize information flow. In our experiment, we show that on average, important information flow through skip connections 3 times more often than attentions. 
    \item By visualizing the extracted patterns, we show how information flow of words interact inside the model and BERT may use grammatically incorrect cues to make predictions. 
    \item The consistency of influence patterns across instances of a task reflects BERT's performance for that task.  
    \item Through ablation experiments, we find that influence patterns account for information flows in BERT on average 74\% and 25\% more accurately than prior attention-based and layer-based explanation methods\cite{abnar2020quantifying, leino2018influence, dhamdhere2018important}, respectively. 
    %Further, for grammatical tasks, such as SVA and RA, grammatically and semantically influential words follow more consistent patterns.

\end{itemize}

\section{Background}\label{sec:background}
We begin this section by introducing notations and architecture of BERT in Sec.~\ref{sec:background-notations}. We then introduce \emph{distributional influence} as an axiomatic method to explain the output behavior of any deep model in Sec.~\ref{sec:background-distributional-influence}, which serves as a building block for our method to follow in the next section.

% Finally, we discuss how \emph{distributional influence} is generalized to explain the contextualization process in LSTM, a neighboring language model to BERT, in Sec.~\ref{sec:background-influence-pathway}.
% \input{figure_BERT}
\input{figure_BERT}

% \begin{figure*}
%     % \centering
%     % \includegraphics[width=0.9\textwidth]{images/BERT_pattern.png}
%     \input{figure_BERT}
%     \caption{BERT Transformer architecture (left) and details of a transformer layer (right) for an instance of the SVA task, which evaluates whether the model chooses the correct verb form \textit{IS} over \textit{ARE} to agree with the subject. An example of a pattern is highlighted with cyan nodes. }
%     % .  An example of a pattern (Def.~\ref{def: mppi}) $\pi = [h^0_m, h^L_m]$ is highlighted with cyan nodes.
%     % Refining $\pi$ with our proposed method, \emph{Guided Patter Refinement}, results in an embedding-level path(highlighted with red edges in the left) and an attention-level path with extra nodes and edges (highlighted in red in the right). }
%     \label{fig:berts}
% \end{figure*}

\subsection{BERT}\label{sec:background-notations}
Throughout the paper we use $\xvec$ to denote a vector and $|S|$ to denote the cardinality of a set $S$ or number of nodes in a graph $S$. We begin with the basics of the BERT architecture~\cite{devlin2018bert, Vaswani2017AttentionIA} (presented in Fig.~\ref{fig:berts}). Let $ L $ be the number of layers in BERT, $ H $ the hidden
dimension of embeddings at each layer, and $ A $ the number of attention heads. The list
of input word embeddings is $\xvec \defeq [\xvec_1, \xvec_2, ..., \xvec_{N}], \xvec_i \in \mathbb{R}^d $.
%, \forall i \in [0, N-1]$
We denote the output of the $l$-th layer as $\hvec^l_{1:N}$. First layer
inputs are $\hvec^0_{1:N} \defeq \xvec_{1:N} $.We use $a^{l, i}_{j}$ to denote the $j$-th attention head from the $i$-th embedding at $l$-th layer and $s^l_j$ to denote the skip connection that is ``copied'' from the input embedding from the previous layer then combined with the attention output. 
The output logits are denoted by $\mathbf{y}$. 
% Probability scores for candidates of \texttt{[MASK]} are denoted by $\mathbf{y}_i \defeq
% \texttt{logsoftmax}(W\hvec^{L}_i), W \in \mathbb{R}^{C\times H}$ where $ C $ is the vocabulary size.
% as the final classification layer and outputs are the predicted word $\hat{y}_i \defeq \arg\max_c
% \mathbf{y}_{i, c}$ for the $i$-th token. 
% We denote the index of \texttt{[MASK]} as $m$. \caleb{replace with actual numbers} 
% This architecture is presented in
% \fref{fig:berts}.

\paragraph{Computation Graph of BERT} A deep network can be viewed as a computational graph $\mathcal{G} \defeq (\mathcal{V},\mathcal{F}, \mathcal{E})$, a set of nodes, activation functions, and edges, respectively. In this paper, we assume the graph is directed, acyclic, and does not contain more than one edge per adjacent pair of
nodes.
% \footnote{The single edge restriction is for notational convenience; if a given
%   neural model does have more than one edge between adjacent nodes, we can replace duplicate edges
%   with 2-length paths through dummy identity nodes to satisfy this requirement without affecting
%   its semantics.}.
  A path $ p $ in $ \mathcal{G} $ is a sequence of graph-adjacent nodes $ [p_1, p_2, \cdots, p_t ] $; $p_t$ is the output node. Thus, the Jacobian passing through a path $ p $ evaluated at input $ \xvec $ is  $
\prod^{-1}_{i=1} \partial p_{i}(\mathbf{x})/\partial p_{i-1}(\mathbf{x})$ as per chain rule. 
We further denote the Jacobian of the output of node $ n_i $ w.r.t the output of connected (not necessarily
directly) predecessor node $ n_j $ evaluated at $\xvec$ as $\partial n_j(\mathbf{x})/\partial
n_i(\mathbf{x})$.  

For computational and interpretability reasons, an ideal graph would contain as few nodes and edges as possible while exposing the structures of interest. For BERT we propose two graphs: \emph{embedding-level graph} $ \mathcal{G}_e $ corresponding to the nodes and edges shown in \fref{fig:berts} (left) to explain how the influence of input embeddings flow from one layer of internal representations to another and to the eventual prediction; and \emph{attention-level graph} $ \mathcal{G}_a \supset \mathcal{G}_e $ that additionally includes attention head nodes and skip connection nodes as in \fref{fig:berts} (right), a finer decomposition to demonstrate how influence from the input embedding flows through the attention block (or skip connections) within each layer.

\subsection{Explaining Deep Neural Networks}\label{sec:background-distributional-influence}

Gradient-based explanations~\cite{sundararajan2017axiomatic,dhamdhere2018important, smilkov2017smoothgrad} are well-studied in explaining the behavior of a deep model by attributing the model's output to each input feature as its feature importance score. While existing approaches often explain with the most important features for a single class $i$, \emph{distributional Influence}~\cite{leino2018influence} generalizes gradient-based approaches to answer a broader set of questions, e.g. \textit{why \texttt{[MASK]} should be \texttt{IS} instead of \texttt{ARE}} (Figure~\ref{fig:berts}), by introducing \emph{quantity of interest} (QoI). Suppose a general network $f:\mathbb{R}^d\rightarrow\mathbb{R}^n$, a QoI is a differentiable function $q(f(x))$ that outputs a scalar to incorporate the subject of an explanation. For example, to answer the aforementioned question, we can define $q(f(x)) = f(x)_{\texttt{IS}} - f(x)_{\texttt{ARE}}$ where $f(x)_*$ is the logit output of class $*$. Formally, we introduce \emph{Distributional Influence}.

\begin{definition}[Distributional Influence]
\label{def: distributional influence}
  Given a model $ f: \mathbb{R}^d \rightarrow \mathbb{R}^n $, an input $ \mathbf{x} $, a user-defined distribution $\mathcal{D}(\mathbf{x}) $, and a user-defined QoI $q$, \emph{Distributional Influence} $ g(\mathbf{x};q, \mathcal{D})  $ is defined as:
$$g(\mathbf{x};q, \mathcal{D}) \defeq
    \E_{\mathbf{z} \sim \mathcal{D}(\xvec)} \frac{\partial q(f(\mathbf{z}))}{\partial \mathbf{z}}$$

\end{definition}

\begin{remark}\label{remark:influence}
% Saliency~\cite{baehrens2010explain}, as an example, defines influence as
% the gradient of output w.r.t. the input. In a generalized framework of \cite{leino2018influence}, influence quantifies the impact of each input feature towards a concept (e.g. SVA) by
% instrumenting a model's inputs with a \emph{distribution of interest} (DoI) and the output with a
% \emph{quantity of interest} (QoI). 
Distributional Influence also leverages a user-defined distribution $\mathcal{D}(\xvec)$ to capture the network's behavior in a neighborhood of the input of interest $\xvec$. By introducing $\mathcal{D}(\xvec)$, we can prove that several popular attribution methods are specific cases of the distributional influence; e.g. when $\mathcal{D}(\xvec)$ is a Gaussian Distribution, $g(\mathbf{x};q, \mathcal{D})$ reduces to Smooth Gradient\cite{smilkov2017smoothgrad}; when $\mathcal{D}(\xvec)$ is a uniform distributions over a path $c = \{\xvec + \alpha(\xvec-\xvec_b), \alpha \in [0, 1]\}$ from a user-defined baseline input $\xvec_b$ to $\xvec$, $\mathcal{D}(\xvec)$ reduces to Integrated Gradient~\cite{smilkov2017smoothgrad}.
\end{remark}

%\citet{acl} define SVA by setting the distribution of interest as the linear interpolation between
%the embedding of a singular noun and the plural noun (so as to exercise a subject's grammatical
%number) and setting the quantity of interest as the difference in probability of the correctly
%numbered verb and the incorrectly numbered verb (we formalize a variant of this concept for BERT's
%\verb|[MASK]| prediction setting and equivalent %

% Instantiations defining SVA and RA concepts in BERT models are found in Sec.~\ref{sec:evaluation}. 

% Examples of DoI include Gaussian distributions with mean $\xvec$~\citep{smilkov2017smoothgrad}, or uniform distributions over a path $c = \{\xvec + \alpha(\xvec-\xvec_b), \alpha \in [0, 1]\}$ from a user-defined baseline input $\xvec_b$ to the target input $\xvec$ %, which is then the line  integral $ \int_{0}^1 \frac{\partial q(f(\xvec))}{\partial \xvec} \paren{c(\alpha)} \cdot p(\alpha)  \; d \alpha $ where $ p $ is the p.d.f. of the DoI
% ~\citep{sundararajan2017axiomatic}. 

We use $\mathcal{D}(\xvec)$ as the uniform distribution over a linear path described in Remark~\ref{remark:influence} in the rest of the paper because it provides several provable properties\cite{sundararajan2017axiomatic} to ensure the faithfulness of our explanations. We approximate the expectation in Def.~\ref{def: distributional influence} by sampling discrete points in the uniform distribution.

\section{Tracing Influence Flow with Patterns}\label{sec:methods}

% Contextualization refers to the concept where a word embedding is sensitive to the context in which it appears~\cite{ethayarajh2019contextual}. For example, to predict which one, \texttt{is} or \texttt{are}, is a better fit for \texttt{[MASK]} in the sentence, \textit{the boys behind the tree \text{\texttt{[MASK]}} running}, the initial embedding of \texttt{[MASK]} needs to absorb information from the context and 

To explain how different concepts in the input flow to final predictions in BERT, it is important to show how the information from each input word flows through each intermediate layer and finally reaches the output embedding of interest, e.g. \texttt{[MASK]} for pretraining or \texttt{[CLS]} for fine-tuned models. Prior approaches have use the attention weights to build directed graphs from one embedding to another~\cite{abnar2020quantifying, wu2020structured}. These approaches use heuristics to treat high attention weights as indicators of important information flow between layers. However, as more work starts to highlight the axiomatic justifications of gradient-based methods over attentions weights as an explanation approach \cite{treviso2020explanation}, we therefore explore an orthogonal direction in applying distributional influence in BERT to trace the information flow. Since distributional influence only attributes the output behavior over the input features, in this section, we generalize it to find important internal components that faithfully account for the output behavior. 
% We begin our discussion by viewing BERT as a computational graph described in Sec.~\ref{sec:background-notations}, which provides a unified way of treating different modules in the network for the ease of analysis in the later sections.

% cannot show if or how they are contextualized internally to form higher-level concepts. \textit{Influence Paths}~\citep{DBLP:conf/acl/LuMLFD20} localizes an input influence measurement to paths in a neural model, and thus can be used to show how the influence of the input representations flows internally through one internal representation to another.

\paragraph{Tracing Influence by Patterns} By viewing BERT as a computation graph ($\mathcal{G}_e$ or $\mathcal{G}_a$ in \ref{sec:background-notations}), we restate the problem: given a source node $s$ and a target node $t$, we seek a significant pattern of nodes from $s$ to $t$ that shows how the influence from $s$ traverses from node to node and finally reaches $t$. An exhaust way to rank all paths by the amount of influence flowing from $s$ to $t$ is possible in smaller networks, as is done by Lu et al.~\cite{DBLP:conf/acl/LuMLFD20}. However, the similar approach lacks scalability to large models like BERT. Therefore, we propose a way to greedily narrow down the searching space from all possible paths to specific \emph{patterns}. That is, our approach is two-fold: 1) we employ abstractions of sets of paths as the localization and influence quantification instrument; 2) we discover influential patterns with a greedy search procedure that refines abstract patterns into more concrete ones, keeping the influence high. We begin with the formal definition of a \emph{pattern}.

\begin{figure*}
    \centering
    \includegraphics[width=.8\textwidth]{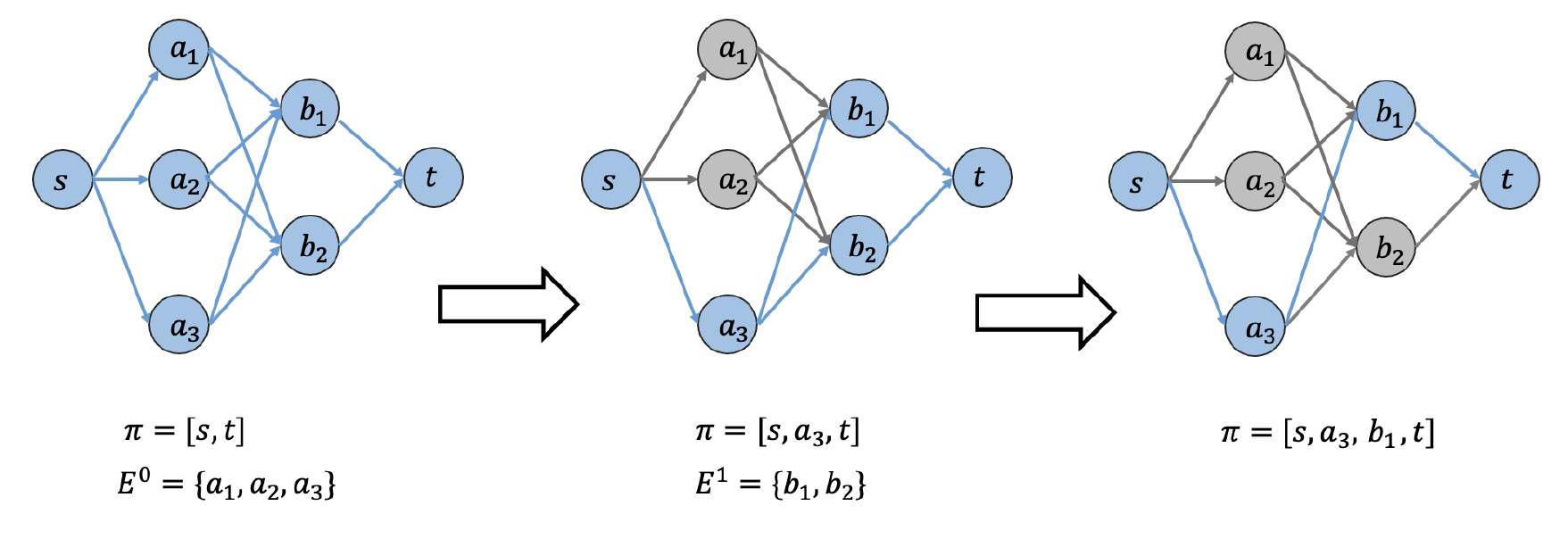}
    \caption{A visual illustration of Guided Pattern Refinement (GPR) in a toy example. We start with a pattern $\pi = [s, t]$ containing only the source and the target node. At each step we define a guided set $E^0$ and $E^1$, respectively and find the node in the guided set that maximize the pattern influence (Def.~\ref{def: mppi}). GPR finally returns a pattern $\pi = [s, a_3, b_1, t]$ that abstracts a single path.}
    \label{fig:example-gpr}
\end{figure*}

\begin{definition}[Pattern]
  \label{def: mpp}
  A pattern $ \pi $ is a sequence of nodes $ [ \pi_1, \pi_2, \cdots, \pi_{-1}]
  $ such that for any pair of nodes $ \pi_i , \pi_{i+1} $ adjacent in the sequence (not necessarily
  adjacent in the graph), there exists a path from $ \pi_i $ and $ \pi_{i+1} $.

\end{definition}
A pattern $\pi$ abstracts a set of paths, written $ \gamma(\pi) $ that follow the given
  sequence of nodes but are free to traverse the graph between those nodes in any way. Interpreting
  paths and patterns as sets, we define $\gamma(\pi) \defeq \set{p \subseteq \mathcal{P} \suchthat
    \pi \subseteq p} $ where $ \mathcal{P} $ is the set of all paths from $ \pi_1 $ to $ \pi_{-1}
  $. If every sequence-adjacent pair of nodes is directly connected then the pattern abstracts a
  single path. To quantify the amount of information that flows from the input node to the target node over a particular pattern, we propose \emph{Pattern Influence}, motivated by distributional influence.  
%   Two propositions follow the Def.~\ref{def: mppi} as its natural consequences, the corresponding proofs of which are left in the Appendix.
  
%   We show Proposition~\ref{prop:pattern-influence} as a direct outcome of Def.~\ref{def: mppi}, followed by Corollary~\ref{Cor:pattern-influence}.

\begin{definition}[Pattern influence]
  \label{def: mppi}
  Given a computation graph and a user-defined distribution $ \mathcal{D} $, the \emph{influence} of an influence pattern $ \pi $, written $ \mathcal{I}(\xvec, \pi) $ is the total influence of all the
  paths abstracted by the pattern: $ \mathcal{I}(\mathbf{x}, \pi)
    \defeq  \sum_{p \in \gamma(\pi)} \mathbb{E}_{\mathbf{z} \sim \mathcal{D}(\xvec)} \prod^{-1}_{i=1} \frac{\partial p_{i}(\mathbf{z})}{\partial p_{i-1}(\mathbf{z})}$.
\end{definition}

\begin{proposition}[Chain Rule]\label{prop:pattern-influence}
 $\mathcal{I}(\mathbf{x}, \pi) = \mathbb{E}_{\mathbf{z} \sim \mathcal{D}(\xvec)}
    \prod_{i=1}^{-1}\frac{\partial \pi_i(\mathbf{z})}{\partial \pi_{i-1}(\mathbf{z})}$ for any distribution $\mathcal{D}(\xvec)$.
\end{proposition}

\newcommand{\graphe}{\mathcal{G}_e}
\newcommand{\grapha}{\mathcal{G}_a}

Prop.~\ref{prop:pattern-influence} (proof in Appendix~\ref{appendix: distributional influence}) simplifies the evaluation of the pattern influence by specifying the exact set of internal nodes through which influence flows in a computational graph. 
% Besides backward propagation after the forward pass, many frameworks, .e.g Tensorflow\cite{tensorflow2015-whitepaper}, have also implemented the forward Jacobian computation to efficiently calculate $\partial \pi_i(\mathbf{z})/\partial \pi_{i-1}(\mathbf{z})$. 
% Leveraging the multiplicative nature of pattern influence, 
% We now introduce a greedy way of finding the most important path between a source node and a targeting by maximizing the pattern influence and refining a pattern into a path.
We hereby introduce a greedy way of finding the most influential pattern in both $\mathcal{G}_e$ and $\mathcal{G}_a$.

\newcommand{\subst}[0]{\; \backslash \;}
\noindent\textbf{Guided Pattern Refinement(GPR)} Starting with source and target nodes $ s $ and $ t $ along with a initialized pattern $ \pi^0 = \set{s,t} $ representing all paths between $ s $ and $ t $, we construct $ \pi^1 $ by adding one\footnote{The algorithm can be easily adapted to include more nodes per layer. However, we found one node per layer retain a reasonable proportion of influence for the tasks evaluated in this paper(See Sec.\ref{sec: gsig}).} sequence-adjacent node $e^0$ (there is a direct path between $s$ and $e^0$) from a \emph{guiding set} $ E^0 $ that maximizes the influence of the resulting pattern such that:
\begin{align}
    % e^0 = \arg\max_{e \in E^0} \mathcal{I}(\xvec, \pi^1) = \arg\max_{e \in E^0} \{\mathcal{I}(\xvec, \sparen{s, e, t})
    e^0 =  \arg\max_{e \in E^0} \{\mathcal{I}(\xvec, \sparen{s, e, t}), 
    \pi^1 = \sparen{s, e^0, t}, 
    % \arg\max_{e \in E^0} \{\mathcal{I}(\xvec, \sparen{s, e^0}) \cdot \mathcal{I}(\xvec, \sparen{e^0, t})\}
\end{align}
This procedure is iterated until we exhaust the last guiding set. We show an example of GPR in a toy graph in Fig.~\ref{fig:example-gpr}. For an embedding-level graph $\mathcal{G}_e$, each guiding set $E^l$ includes all embeddings $\hvec^l_{1:N}$ at layer $l$. 
% At the first iteration and subsequently, the guiding set defines a cut of the (multi-partite) graph between two sequence adjacent nodes (initially just $ s $ and $ t $). 
% The procedure is repeated until the last guiding set $E^L$ where the embedding-level pattern $\pi^e$ returned by GPR contains one internal embedding per layer in $\mathcal{G}_e$. 
For the attention-level graph $\mathcal{G}_a$, we refine on the embedding-level pattern $\pi^e$ by only expanding $\pi^e$ from addition nodes in $\mathcal{G}_a$ compared with $\mathcal{G}_e$: we perform GPR iterations between $\pi^e_l, \pi^e_{l+1} \in \pi^e$ with the guiding set $E^l_a$ which includes $A$ attention heads $\mathbf{a}^l$ and the skip node $\mathbf{s}^l$(Fig.~\ref{fig:berts} Right), until we reach the same last node in $\pi^e$. The returned attention-level pattern $\pi^a$ thus abstracts a single path from the source to the target in $\mathcal{G}_a$. As the
attention-level analysis refines the embedding-level analysis, the produced attention-level pattern
$ \pi^a $ abstracts a strict subset of the paths of the attention-level graph that the
embedding-level pattern $ \pi^e $ abstracts. That is, $ \pi^e \subset \pi^a $ while $ \gamma(\pi^a) \subset \gamma(\pi^e)$. The detailed algorithm of GPR and analysis of its optimality can be found in Appendix~\ref{appendix: pseudo-code} and \ref{appendix: optim-GPR}.

\section{Experiment}\label{sec:evaluation}

Experiments in this section demonstrate our method as a tool for interpreting end-to-end information flow in BERT. Specific visualizations exemplify these interpretations including the importance of skip connections and BERT's encoding of grammatical concepts are included in \sref{sec: path visualization}. \sref{sec:entropy} explores the consistency of patterns across instances and template positions and how they relate to task performances and influence magnitudes in.  Sec~\ref{sec: gsig}  demonstrates two advantages of patterns in explaining the information flow of BERT over baselines: 1) abstracted patterns, with much fewer nodes compared to the whole model, carry sufficient information for the prediction. That is, without the information outside the refined pattern, the model shows no significant performance drop; 2) At the same time, patterns are sparse but concentrated in BERT's components.

% By comparing with several baseline methods we show the pattern's capability of capturing model performance in Sec.~\ref{sec: gsig}. We further show that refined patterns are sparse but concentrated in BERT's components. 

%This section focuses on evaluating the proposed technique against existing baseline methods to demonstrate that influential patterns provide faithful explanations to the internal contextualization of embeddings in BERT. We firstly introduce the setup of our experiments and the baseline methods. We secondly provide quantitative comparisons in Sec.~\ref{sec: gsig}. We end this section with a closer look at how patterns variability relates to task performance.}

\subsection{Setup}\label{sec:setup}

\noindent\textbf{Tasks.} We consider two groups of NLP tasks: (1) \emph{subject-word agreement (SVA)} and
\emph{reflexive anaphora (RA)}. We explore different forms of sentence stimuli(subtask) within each task: object
relative clause (Obj.), subject relative clause (Subj.), within sentence complement (WSC), and
across prepositional phrase (APP) in SVA~\citep{marvin2018targeted}; number agreement (NA) and gender agreement (GA) in RA~\citep{lin2019open}. Both datasets are evaluated using masked language model (MLM) as is used in Goldberg~\cite{goldberg2019assessing}. We sample 1000 sentences from each subtask evenly distributed across different sentence
types (e.g. singular/plural subject \& singular/plural intervening noun) with a fixed sentence
structure;  (2) \emph{sentiment analysis(SA)}: we use 220 short examples (sentence length$\leq 17$) from the evaluation set of the 2-class GLUE SST-2 sentiment analysis dataset \cite{wang2018glue}. More details and examples of each task can be found in Appendix~\ref{appendix: Linguistic tasks}.
%   sentence length and the word types in each position are
% consistent across samples.
%   \todo{caleb: expand experiments to include more data}

\noindent\textbf{Models.}
For linguistic tasks, We evaluate our methods with a pretrained BERT($L=6, A=8$)~\cite{turc2019}, referred hereby as $\text{BERT}_{\text{SMALL}}$. For SST-2 we fine-tuned on the pretrained $\text{BERT}_{\text{BASE}}$\cite{devlin2018bert} with $L=12, A=12$. The models are similar in sizes compared to the transformer models used in \cite{abnar2020quantifying}. All computations are done with a Titan V on a machine with 64 GB of RAM.
% and has comparable performance to larger models such as $\text{BERT}_{\text{BASE}}$ in the evaluated tasks. 
See Appendix~\ref{appendix: setup} for more details. 

% \noindent\textbf{QoI and Distributional Influence.} We use QoI defined in Sec.~\ref{sec:background-distributional-influence}\cite{leino2018influence, DBLP:conf/acl/LuMLFD20} where $q(\yvec_m) \defeq
% y_{m,correct} - y_{m, wrong}$, e.g. $y_{m,\texttt{IS}}-y_{m,\texttt{ARE}}$ for \texttt{she [MASK] happy} in the SVA task and $y_{m,\texttt{positive}}-y_{m,\texttt{negative}}$ for \texttt{The movie is good.} in sentiment analysis task. 

% We select $\mathcal{D}$ as an uniform distribution over a linear path from $\xvec_b$ to
% $\xvec$ in the input space for each word with the baseline $\xvec_b$ defined as the
% the input embedding of \texttt{[MASK]}, which can be viewed as a neutral word with no information.

\noindent\textbf{Implementation of GPR.} Let the target node for SVA and RA tasks be the output of the QoI score $q(\yvec) \defeq y_{correct} - y_{wrong}$. For instance, $y_{\texttt{IS}}-y_{\texttt{ARE}}$ for the sentence \texttt{she [MASK] happy}. Similarly, we use $ y_{\texttt{positive}}-y_{\texttt{negative}}$ for sentiment analysis. We choose an uniform distribution over a linear path from $\xvec_b$ to
$\xvec$ as the distribution $\mathcal{D}$ in Def.~\ref{def: mppi} where the $\xvec_b$ is chosen as the
the input embedding of \texttt{[MASK]} because it can viewed a word with no information. For a given input token $\xvec_i$, we apply GPR differently depending on the sign of distributional influence $g(\mathbf{x}; q, \mathcal{D})$: if $g(\mathbf{x}; q, \mathcal{D})\geq 0$, we maximize the pattern influence towards $q(\yvec)$ at each iteration of the GPR otherwise we maximize pattern influence towards $-q(\yvec)$. We use $\inputpattern$ as the extracted patterns for individual input word $i$. When explaining the whole input sentence, we collect all refined patterns for each word and use $\allpatterns = \bigcup_i \pi_i$. Further, we denote $\positivepatterns$ as the set of patterns for all positively influential words. Both terms may be further decorated by $ a $ or $ e $ to denote attention-level or embedding-level results.

% Note the last two baselines are layer-based analysis which can neither isolate the flow of information from each word, nor isolate skip connections, therefore not always directly comparable with our method.
% (3) Patterns derived from layer-based attribution methods including $\baselineinfluencepatterns$ from influence-directed explanations\cite{leino2018influence} and $\baselineconductancepatterns$ from conductance\cite{dhamdhere2018important}. Since these methods only compute the influence(attribution) of nodes(neurons) in each layer without traversing the whole graph, to directly compare with our method, we define baseline patterns by choosing the most influential nodes from each layer computed from each layer-based method. However, layer-based methods cannot inspect the flow of information from each word, whic ... \todo{fill in} 

% More implementation details for these baselines can be found in Appendix~\ref{appendix: Attention Path}.

\subsection{Visualizing Influence Patterns} \label{sec: path visualization}

\begin{figure*}[t]
  \begin{subfigure}[]{.33\textwidth}
    \centering
    \includegraphics[width=\linewidth]{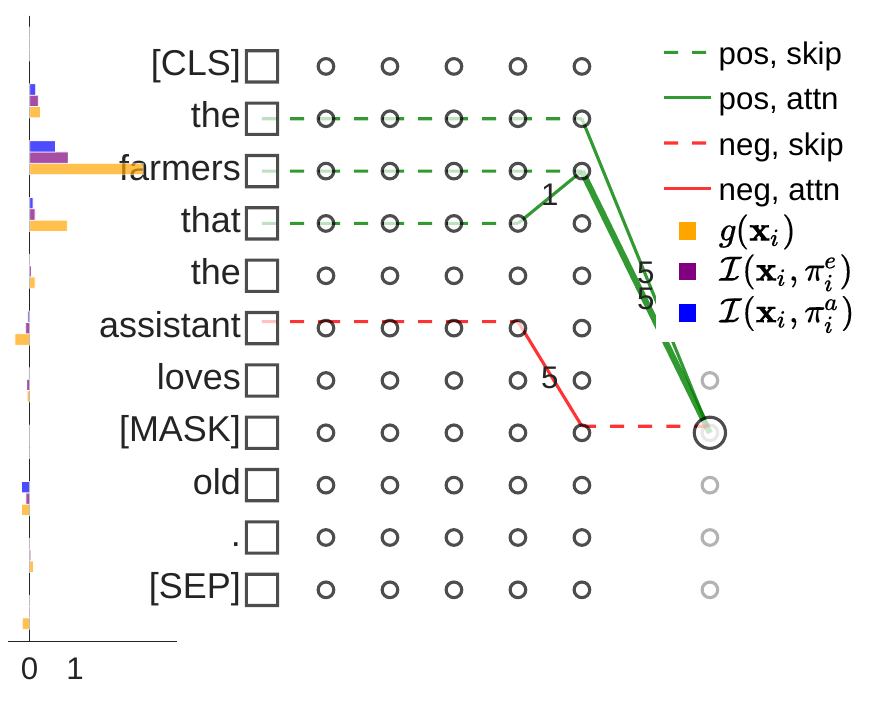}
    \caption{plural subject+singular attractor(PS)}
    \label{fig:gpr_ps}
  \end{subfigure}
%   \hfill
  \begin{subfigure}[]{.33 \textwidth}
    \centering
    \includegraphics[width=\linewidth]{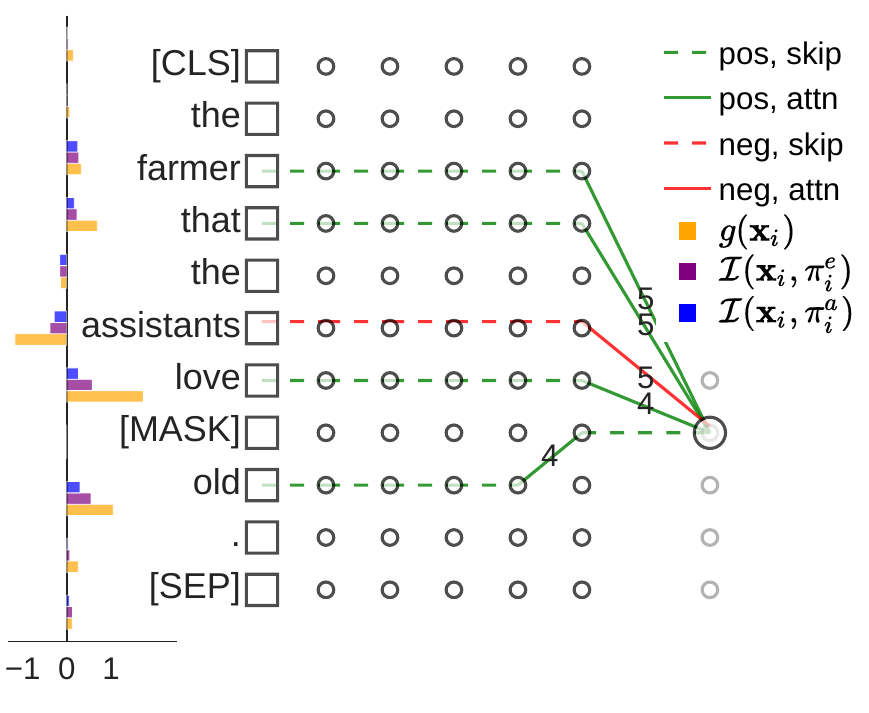}
    \caption{singular subject+plural attractor(SP)}
    \label{fig:gpr_sp}
  \end{subfigure}
  \begin{subfigure}[]{.33 \textwidth}
    \centering
    \includegraphics[width=\linewidth]{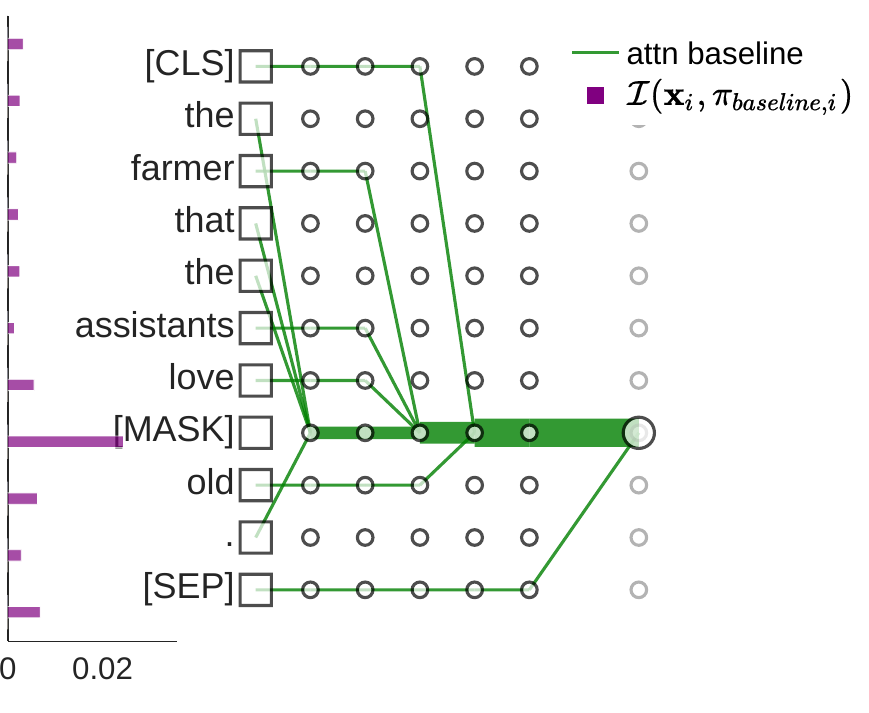}
    \caption{attention-based baseline $\baselineattentionpatterns$}
    \label{fig:gpr_sp_rollout}
  \end{subfigure}
  %   \medskip
  \caption{(a)(b) Patterns for two instances of
    \textit{SVA-Obj}.(c)Baseline pattern $\baselineattentionpatterns$. For each plot: Left: bar plots of the
    distributional influence $g(\xvec_i)$(yellow), $\mathcal{I}(\xvec_i, \pi^e_i)$ (purple) and $\mathcal{I}(\xvec_i, \pi^a_i)$
    (blue) (or $\mathcal{I}(\xvec_i, \pi_{baseline,i})$) for each word at position $ i $. Right: Extracted patterns $\Pi$ from selective words. Square nodes and circle nodes denote input and internal embeddings, respectively. In (a) and (b), influence flowing through skip connections is represented by dashed lines and
    attention heads in solid lines; the edges are are marked with the corresponding
    attention head number (ranging from 1 to $A$) in  $\Pi^a$ . Line colors represent the sign of influence (red as negative and green as positive).}
  \label{fig:intro}
%   \todo{fix legend}
%   \caleb{One reviewer comments on this figure being here too early. Wondering if we can better use this figure to convey the merits of our approach in intro}\zifan{I think, we should replacing Fig 1 with a simplified version of it, by taking out concentration plot on the left side, which does make sense to be discussed so early.}
\end{figure*}

This section does not serve as an evaluation but an exploration of insightful results and succinct conclusions an human user can learn from our proposed technique. We visualize the information flow identified by patterns found by GPR, and compare with those generated by attention weights as explored in the literature~\cite{abnar2020quantifying, wu2020structured}\footnote{The two works compute attribution from input to internal embeddings by aggregating attention weights(averaged across heads) across layers while attribution to individual paths is implicit.}. We therefore use $\baselineattentionpatterns$ to denote a pattern of nodes by maximizing the product of the corresponding attention weights between each pairs of nodes from adjacent layers. Implementation details are included in Appendix~\ref{appendix: Attention Path}.

% \noindent\textbf{$\blacksquare$ $\baselineattentionpatterns$}: Inspired\footnote{The two works compute attribution from input to internal embeddings by aggregating attention weights(averaged across heads) across layers while attribution to individual paths is implicit.} by \cite{abnar2020quantifying} and \cite{wu2020structured}, we find significant patterns from the source to the target by maximizing the product of the corresponding attention weights between each pairs of nodes from adjacent layers.

We first focus on instances of the subtask \textit{SVA across object relative clauses (SVA-obj)}, which are generated from the template: \texttt{the SUBJECT that the ATTRACTOR VERB [MASK](is/are) ADJ.}. We observe in Figure~\ref{fig:intro} that the subject words exert positive input influences on the correct choice of the verb, and the intervening noun (attractor) exerts negative influence, which is true for both $\mathcal{I}(\xvec_i, \pi^e_i)$ and $\mathcal{I}(\xvec_i, \pi^a_i)$ (blue and purple bars in Figure~\ref{fig:gpr_ps} and~\ref{fig:gpr_sp}). While $\baselineattentionpatterns$(Fig.~\ref{fig:gpr_sp_rollout}) does not distinguish between positively influential words and negative ones, nor do they show an interpretable pattern. Our other main findings are discussed as follows.

\paragraph{Finding I: Skip Connection Matters.} Horizontal dashed lines in Figure~\ref{fig:intro} indicate that influence can flow through layers at the same word
position via skip connections, which is not isolated as separate nodes $\baselineattentionpatterns$ (shown in Fig.~\ref{fig:gpr_sp_rollout}). Fig.~\ref{fig:gpr_ps} and~\ref{fig:gpr_sp} also show that the influence from subject words \texttt{farmer[s]} travels
through skip connections across layers before it transfers into the attention head 5 in the last
layer, indicating the ``number information'' from the
subject embedding flows directly to the output.
% , however, exactly how it overcomes(or not)
% the comparable signals from the attractors is explained the next paragraph.
% \caleb{not sure if this connection should be made here or in the related work, but I think it is a very important connection to make}
 Interestingly, this also explains why attentions can be
  pruned effectively without compromising the performance~\cite{kovaleva2019revealing,michel2019sixteen,voita2019analyzing} as important information may not flow through
  attentions at all. Namely, they are simply ``copied'' to the next layer through the skip
  connection. In fact, attention heads are traversed far less often than skip connections, which account for $75.4\%$ of nodes in $\Pi_a$ across all tasks evaluated. 
 
\begin{figure}[t]
\centering
\begin{subfigure}[]{.42\textwidth}
    \centering
    \includegraphics[width=\linewidth]{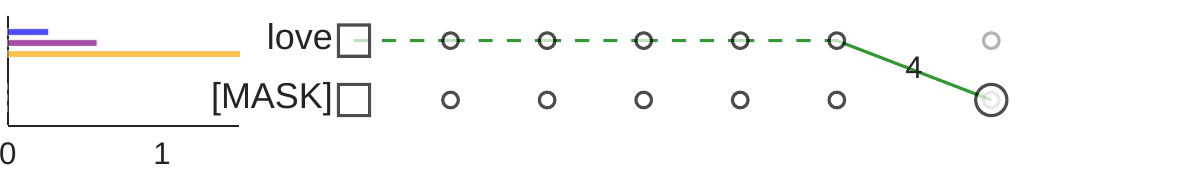}
    % \vfill
    \includegraphics[width=\linewidth]{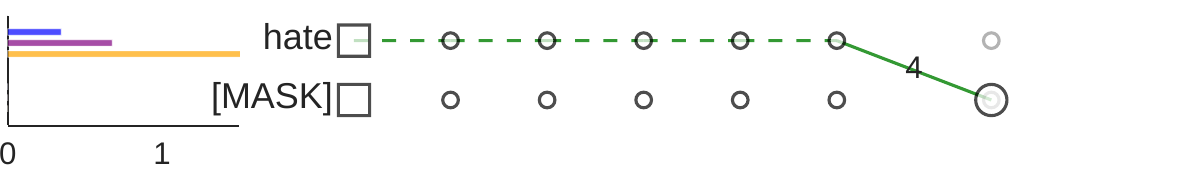}
    % \vfill
    \includegraphics[width=\linewidth]{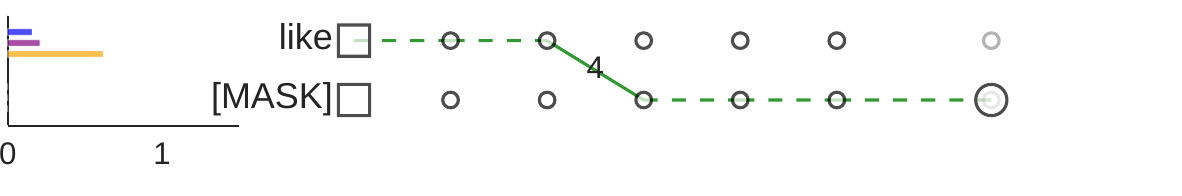}
    \caption{Interactions between clausal verbs \& \texttt{[MASK]} for \textit{SVA-Obj}}
    \label{fig:clausal_verb}
\end{subfigure}
\begin{subfigure}[]{.55\textwidth}
    \centering
    \includegraphics[width=\linewidth]{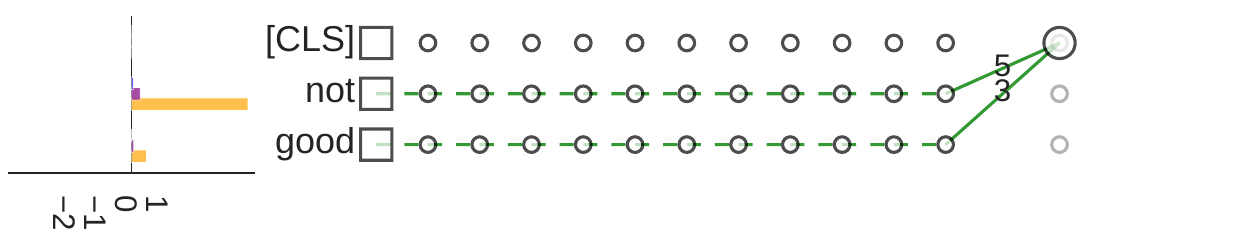}
    \vfill
    \includegraphics[width=\linewidth]{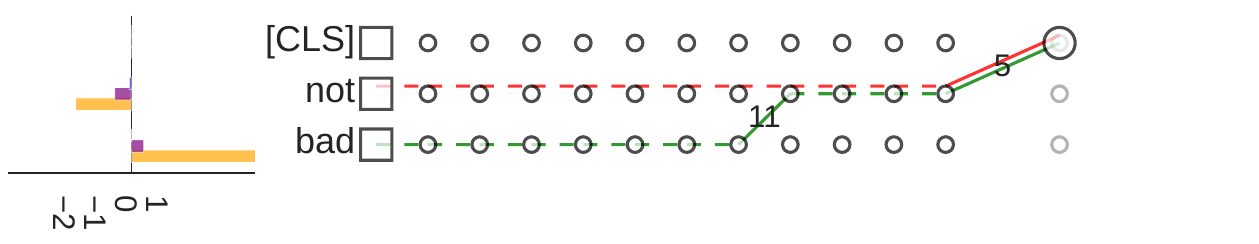}
    \caption{Interactions between \textit{not} \& adjectives in SA}
    \label{fig:sst_not}
\end{subfigure}
\caption{(a) Patterns on three clausal verbs from SP, \textit{SVA-obj}.(b) Patterns for two instances in the SA task. Legend follows Figure~\ref{fig:intro}. }
\label{fig:clausal_verb_not}
\end{figure}

\paragraph{Finding II: \texttt{that} as an Attractor.} Fig.~\ref{fig:gpr_ps} and \ref{fig:gpr_sp} shows that the singular subject is less influential than the plural
subject, especially when compared to the large negative influence from the attractor. Besides, the word \texttt{that} behaves like a
singular pronoun in the singular subject case (Figure~\ref{fig:gpr_sp}), flowing through the same pattern as the subject (skip connections
+ attention head 5), whereas \texttt{that} is more like a grammatical
marker (relativizer) in the plural subject case (Figure~\ref{fig:gpr_ps}): the pattern from \texttt{that} converge to the subject in the second to last layer via a different attention head 1. An explanation to the observed difference is that \texttt{that} can either be used as a singular pronoun in English, or a marker that encodes the
syntactic boundary of the clause to help identify the subject and ignore the
attractor. This finding reveals that although the latter one is always the correct usage for \texttt{that} across all instances, BERT may resort to the ``easier''(while ungrammatical) encoding of \texttt{that} as a pronoun when the number happens to be singular.
% This finding reveals that differentiating the actual use case of \texttt{that} remains a challenge for BERT.

% The key difference is that the word \texttt{that} behaves
% differently for singular and plural subject. \texttt{that} in Figure~\ref{fig:gpr_sp} behaves as a
% singular pronoun (since \texttt{that} also means a singular pronoun in English), flowing through the same pattern as the subject (skip connections
% + attention head 5); \texttt{that} in Figure~\ref{fig:gpr_ps}, however, behaves more like a grammatical
% marker (relativizer): the pattern
% from \texttt{that} converge to the subject in the second to last layer via a different attention head (1). We speculate that \texttt{that} for the plural subject encodes the
% syntactic boundary of the clause to help identify the subject and ignore the
% attractor.
\paragraph{Finding III: \texttt{love}, \texttt{hate} and \texttt{like}}
Figure~\ref{fig:clausal_verb} shows the pattern across three instances containing different clausal verbs by replacing \texttt{love} in Figure~\ref{fig:gpr_sp} with \texttt{like} or \texttt{hate}. We observe that \texttt{hate} and \texttt{love} are more influential, with a distinct and more concentrated pattern compared to that of the word \texttt{like}, while one comparable to that of the subject or noun attractor. Therefore, \texttt{hate} and \texttt{love} are treated more like singular nouns than \texttt{like}, contributing positively to the correct prediction of the verb's number (but for the wrong grammatical reason). This discrepancy also corroborates different accuracy for instances in the three subsets: \texttt{hate}: .99;  \texttt{love}: 1.0;  \texttt{like}: .82. In fact, the number of misclassifications containing \texttt{like} accounts for more than $33\%$ of all misclassifications in \textit{SVA-Obj(SP)}.

\noindent\textbf{Finding IV: Interactions with \texttt{not}. } Two instances of sentiment analysis which BERT classifies correctly (negative for \texttt{not good.} and positive for \texttt{not bad.}) in Fig.~\ref{fig:sst_not} shows that: 1) in \texttt{not good.}, \texttt{not} and \texttt{good} does not interact internally and \texttt{not} carries much more positive influence than \texttt{good} towards the correct class (negative sentiment); 2) in \texttt{not bad.}, on the contrary, \texttt{not} contributes negatively to the correctly class (positive sentiment) and \texttt{bad} interacts with the \texttt{not} in an internal layer and exhibit a large positive influence. This disparity shows the model may treat \texttt{not} as a word with negative sentiment by default, and only when the subsequent adjective is negative, can it be used as a negation marker to encode the correct sentiment. \todo{citation}

% In Figure \ref{fig:sst_not}, we look at instances of the sentiment analysis task fine-tuned with the larger $\text{BERT}_{\text{BASE}}$ model.  In particular, we look at two simple negation examples, both of which BERT classify correctly (negative for \texttt{not good.} and positive for \texttt{not bad.}). However, in the pattern of \texttt{not good.}, there are no internal interactions through attention heads (until the final layer to relay the information to \texttt{[CLS]}). Both \texttt{not} and \texttt{good} contribute positively to the correct class (negative), with the former dominating in influence and \texttt{good} having almost no influence. In the pattern of \texttt{not bad.}, on the contrary, \texttt{not} contributes negatively to the correctly class(positive), while \texttt{bad} interacts with the \texttt{not} in an internal layer and exhibit a large positive influence. The disparity in these two cases shows the model may treat \texttt{not} as a word with negative sentiment by default, and only when the subsequent adjective is negative, can be used as a negation marker to encode the correct sentiment. 

The asymmetries and oddities of patterns of Finding II to IV show how BERT may use ungrammatical correlations for its predictions. We include more examples of patterns in Appendix~\ref{appendix: more viz}.
% \todo{Caleb: one sentence explaining why BERT is better than RNN}

\begin{figure}[t]
\centering
\begin{subfigure}[]{.4\textwidth}
    \centering
    \includegraphics[width = \linewidth]{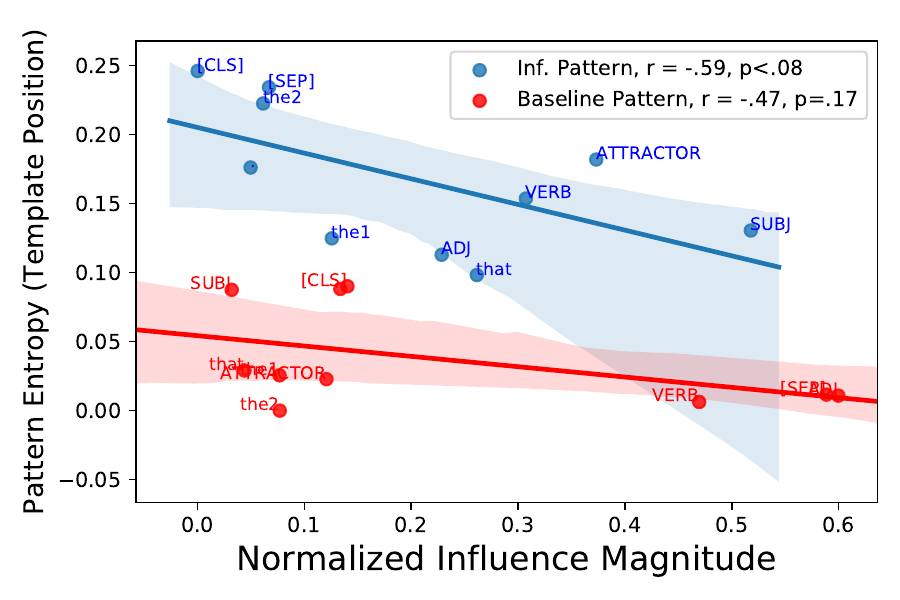}
    \caption{Pattern Entropy vs. Average Influence Magnitude for template positions in \textit{SVA-Obj}}
    \label{fig:variability_position}
\end{subfigure}
\quad
\quad
\begin{subfigure}[]{.4\textwidth}
    \centering
    \includegraphics[width = \linewidth]{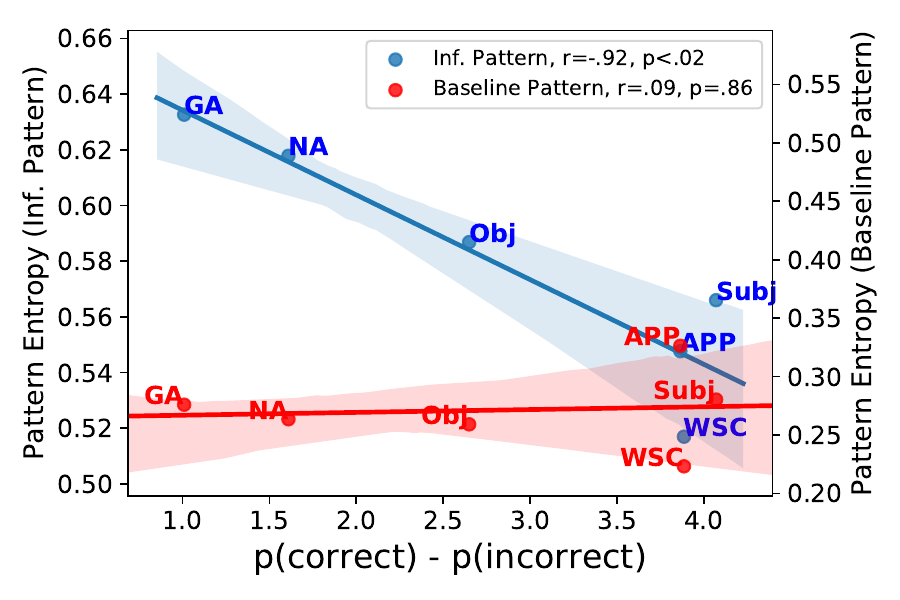}
    \caption{Pattern Entropy vs. Average QoI for linguistic tasks (SVA \& RA)}
    \label{fig:variability}
\end{subfigure}

% \hfill
% \begin{subfigure}[]{.30\textwidth}
%     \centering
%     \includegraphics[width = \linewidth]{images/fig3_ent_group.pdf}
%     \caption{Pattern Entropy vs. Average Influence Magnitude for groups of tokens in Sentiment Analysis }
%     \label{fig:variability}
% \end{subfigure}
\caption{Relationship between Task Performance, Influence Magnitude and Pattern Entropy}
\end{figure}

\subsection{Consistency of Patterns Across Instances}\label{sec:entropy}
%Compressed accuracy shows that patterns extracted for each individual data point faithfully abstract BERT's handling of various tasks, and concentration show that those abstractions are succinct.
The prior section offered examples building connections between information flow and patterns. Since patterns are computed for each word of each instance individually, a global analysis of consistency/variation of patterns across instances can help us make deeper insights and make more general conjectures.
% We observe variation in patterns even on the level of subtask; the patterns for the PS variant of SVA in Figure~\ref{fig:gpr_sp} are at many positions different than those for the SP variant in Figure~\ref{fig:gpr_ps}, despite being the "same" grammatical task. Variation exists also within each of these subtasks: sentences with the same exact grammatical structure may produce different patterns. 
We quantify this variation with \emph{pattern entropy} on the linguistic tasks (SVA \& RA) with fixed templates, where for each instance, each template position are chosen from a group of words. Interpreting the collection of instances, or template positions, in a task as a distribution, pattern entropy is the entropy of the binary probability that a node is part of a pattern, averaged over all nodes (minimally this is 0 if all patterns incorporate the exact same set of nodes). 
With a node $n$ in graph $G$, a set of $K$ instances, $\pi_k$ as the pattern of instance $k$ by abuse of notation and $H$ as the entropy function: pattern entropy of a pattern $\pi$ is defined as: 
$$\hat{H}(\pi, G) = \frac{\sum_{n\in G} H(p_n(\pi))}{|G|}; p_n(\pi) = \frac{\sum_k \mathbbm{1}(n\in \pi_k)}{K}$$

%In Sec.~\ref{sec: path visualization}, we partially answer that question by qualitatively showing that BERT gets confused by various attractors in \textit{SVA-Obj} thus patterns vary even by changing the clausal verb (Figure~\ref{fig:clausal_verb}), indicating that BERT probably does not learn the concept as well as the high task accuracy suggests. 
%In this section, we compare the entropy of patterns $\positivepatterns^e$, by averaging the entropy of each node across all examples, with average QoI, a more accurate quantifier of the task performance than the accuracy. 

Figure~\ref{fig:variability_position} shows an inverse linear relation between influence magnitude and the pattern entropy of each template position($\pi_i$) in \textit{SVA-Obj}. For example, the subject position has on average high influence magnitude and low pattern entropy, while grammatically unimportant template positions such as \texttt{[CLS]} and the period token has relatively low influence and high entropy. Note that this fit is not perfect: attractors, for example,  have high influence magnitude but also high entropy's due to their disparate behaviors among instances shown in Sec~\ref{sec: path visualization}. 

Figure~\ref{fig:variability} demonstrates a more global inverse relation between the pattern entropy of $\positivepatterns$ and task-specific performance(average QoI) for 6 linguistic subtasks studied. This indicates that the more consistent a pattern is for a concept, i.e, the more consistent how a model locates the positively influential signals, the better the model is at capturing that concept. In both figures, the baseline attention-based patterns ($\baselineattentionpatterns$) does not indicate this relation.
% (A more intuitive visualization of pattern entropy in different subtasks can be found in Appendix~\ref{appendix: influence graphs}.)
% We conjecture that this variation is also indicative of performance for these tasks. Whatever is the correct information flow, variation may be a sign that a model is employing some amount of heuristic inferences at expense of sparse generalized rules, as is shown in Sec~\ref{sec: path visualization}. 
Together with Finding II to IV in Sec.~\ref{sec: path visualization}, it demonstrates that patterns may offer alternative insights in formalizing and answering high-level questions such as to what extent does BERT generalize to correct and consistent grammatical rules or use spurious correlations which varies across instances\cite{mccoy2019right, mccoy2020berts, ribeiro2020beyond}.

% However, are patterns among all data points consistent enough to indicate whether BERT actually learns a concept, like humans do? 

% \input{mc_on _embedding_graph}

%\begin{table}[t]
%	\centering
%	\label{tab:qual_data}
%\end{table}

\subsection{Evaluating Influence Patterns }\label{sec: gsig}

This section serves as a formal evaluation of our proposed methods against existing baselines in tracing information flows. Except the attention baseline $\baselineattentionpatterns$ defined in Sec.~\ref{sec: path visualization}, we also include the following baselines (implementation details to follow in Appendix~\ref{appendix: Attention Path}):

\noindent\textbf{$\blacksquare$ $\randompatterns$}: patterns with randomly sampled nodes.

\noindent\textbf{$\blacksquare$ $\baselineinfluencepatterns$}:
patterns consisting of nodes that maximize the \emph{internal influence}~\cite{leino2018influence} for each guiding set.

\noindent\textbf{$\blacksquare$ $\baselineconductancepatterns$}:
patterns consisting of nodes that maximize the \emph{conductance}~\cite{dhamdhere2018important} for each guiding set.

We compare various aspects of extracted patterns against the following baselines, including: (1) ablation experiments showing how influence patterns account for model performance; (2) the sparsity of the patterns in the computational graph. Finally, we also discuss the relation between patterns and attention-weights.
\setlength{\tabcolsep}{2pt}
\renewcommand{\arraystretch}{1}

% \begin{table}[!t]
% \centering
% \resizebox{0.45\linewidth}{!}{%
%     \begin{tabular}{c|cccc|cc}
%     \hline
%      \multirow{2}{*}{Task} & \multicolumn{4}{c}{SVA}& \multicolumn{2}{c}{RA}\\
%      \cline{2-7}
%      & Obj. & Subj. & WSC & APP &  NA & GA \\
%       \hline
%      acc.($\Pi^e_{+}$) &\textbf{.78} &\textbf{.80} &\textbf{1.0} &\textbf{.91} &\textbf{.80} &\textbf{.78} \\
%      acc.($\Pi^e_{r}$) &.56 &.58 &.52 &.52 &.56 &.57 \\
%      acc.($\baselinepatterns$) &.52&.50 &.53 &.61 &.52 &.68  \\
%     \hline
%      acc.($\Pi^a_{+}$) &\textbf{.72} &\textbf{.58} &\textbf{.66} &\textbf{.66} &\textbf{.76} &\textbf{.69} \\
%      acc.($\Pi^a_{r}$) &.52 &.50 &.49 &.49 &.46 &.48 \\
%     %  acc.($\pi^a_{baseline,i}$) & & & & & & \\
%     \hline
%     acc.(ori.) &.96 &1.0 &1.0 &1.0 &.83 &.73 \\
%     \end{tabular}}
%     \caption{With $*\in \{e,a\}$, denoting embedding \& attention-level metrics; acc.($\Pi^*_{+}$), acc.($\Pi^*_{r}$), acc.($\baselinepatterns$) are compressed accuracy of the the positive patterns and two baseline patterns, respectively. \todo{Replace some symbols with words to better corefer to results and their sections in the discussion.}}
% \label{tab: model compressions}
% \end{table}

% \begin{minipage}[c]{0.5\textwidth}
% \centering
\begin{table}[t]
\resizebox{0.45\linewidth}{!}{%
    \centering
    \begin{tabular}{c|cccc|cc|c}
    \hline
     \multirow{2}{*}{Patterns} & \multicolumn{4}{c|}{SVA}& \multicolumn{2}{c|}{RA} & \multirow{2}{*}{SA}\\
     \cline{2-7}
     & Obj. & Subj. & WSC & APP &  NA & GA \\
      \hline
     $\Pi^e_{+}$(ours)&\textbf{.96} &\textbf{.99} &\textbf{1.0} &\textbf{.96} &\textbf{.98} &\textbf{.99} & \textbf{.97}\\
     $\Pi^e_{\text{rand}}$ &.55 &.60 &.55 &.55 &.55 &.56 & .52\\
     $\Pi^e_{\text{attn}}$ &.61&.55 &.49 &.63 &.56 &.65 &  .47\\
     $\Pi^e_{\text{cond}}$ &.93&.91 &.88 &.90 &.71 &.95 &  .94\\
     $\Pi^e_{\text{inf} }$ &.66&.71 &.50 &.56 &.68 &.50 &  .49\\

    \hline
     $\Pi^a_{+}$(ours) &\textbf{.70} &\textbf{.62} &\textbf{.56} &\textbf{.65} &\textbf{.78} &\textbf{.89} & \textbf{.86}\\
     $\Pi^a_{\text{rand}}$ &.50 &.50 &.50 &.50 &.50 &.50 & .49\\
     \hline
     $\Pi^a_{\text{repl\_skip}}$ &.50 &.58 &.54 &.52 &.55 &.50 & .48\\
    %  acc.($\pi^a_{baseline,i}$) & & & & & & \\
    \hline
    original &.96 &1.0 &1.0 &1.0 &.83 &.73 & .92\\
    \hline
    \end{tabular}}
    \quad
\resizebox{0.45\linewidth}{!}{%
    \centering
    \begin{tabular}{c|cccc|cc|c}
    \hline
     \multirow{2}{*}{Metrics} & \multicolumn{4}{c|}{SVA}& \multicolumn{2}{c|}{RA}& \multirow{2}{*}{SA}\\
     \cline{2-7}
     & Obj. & Subj. & WSC & APP &  NA & GA \\
      \hline
     conc.(${\positivepatterns^e}$) & .33& .30 &.27 &.33 &.31 &.22 & .07\\
     conc.(${\negativepatterns^e}$) &.29 &.30 &.38 &.31 &.31 &.23 & .05\\
     share(${\positivepatterns^e}$) & .06 &.09 &.03 &.06 &.04 &.02 & .01\\
    %  $|\gamma_a(\positivepatterns^e)|/|\mathcal{P}^e|$ & .06 &.09 &.03 &.06 &.04 &.02 \\
    \hline
     conc.(${\positivepatterns^a}$) &.21 &.18 &.16 &.18 &.18 &.14 & .01\\
     conc.(${\negativepatterns^a}$) &.17 &.16 &.25 &.15 &.17 &.15 & .01\\
     share(${\positivepatterns^a}$) &\multirow{2}{*}{1.8} &\multirow{2}{*}{1.7}&\multirow{2}{*}{1.6} &\multirow{2}{*}{1.5} &\multirow{2}{*}{1.3} &\multirow{2}{*}{1.3} & \multirow{2}{*}{3e-2} \\
     $(\cdot 10^{-6})$ & & & & & &  \\ 
    %  \hline
     \hline
    %  conc.(${\negativepatterns^a}$) &.17 &.16 &.25 &.15 &.17 &.15 & .01\\
     Alignment &\multirow{2}{*}{.58} &\multirow{2}{*}{.57} &\multirow{2}{*}{.57} & \multirow{2}{*}{.59}&\multirow{2}{*}{.46} &\multirow{2}{*}{.53} & \multirow{2}{*}{.34} \\
     Rate & & & & & & &  \\ 

     \hline
    \end{tabular}}
    % \captionof{table}{share related metrics: conc.(${\positivepatterns^*}$), conc.(${\negativepatterns^*}$): positive/negative concentrations; share(${\allpatterns^*}$): proportion of paths contained in the abstracted patterns over the total number of paths in the corresponding graph.}
% \label{tab:emb_qual}
    % \captionof{table} {}
    % \todo{Replace some symbols with words to better corefer to results and their sections in the discussion.}}

\caption{Summary of quantitative results, with $*\in \{e,a\}$, denoting embedding \& attention-level metrics. Left: Ablated accuracies of $\positivepatterns$ and baseline patterns. Right: Sparsity metrics; conc.(${\pmpatterns^*}$): positive/negative \textit{concentration}; share(${\allpatterns^*}$): \textit{path share} of pattern ${\allpatterns^*}$. }
\label{tab: model compressions}
% \vspace{-6mm}
\end{table}

\noindent\textbf{Ablation.} We extend the commonly-used ablation study for evaluating explanation methods~\citep{deyoung2020eraser, Dabkowski2017RealTI, ancona2018towards, leino2018influence, DBLP:conf/acl/LuMLFD20} from input features to internal patterns as a sanity check of our method. That is, we ablate BERT down to a simpler model: we only retain the nodes from $\positivepatterns^e$ (or $\positivepatterns^a$) while replacing other nodes by zeros. The retained and replaced nodes together are forward passed to the next layer with the original model parameters until a new set of nodes are retained or replaced. \emph{Ablated Accuracy} denotes the accuracy of the ablated model. We show the results for our methods $\Pi^*_{+}$ and other baseline patterns\footnote{we run $\randompatterns$ 50 times per task and average the ablated accuracy with no significant variation, the detailed statistics can be found in Appendix \ref{appendix:random}} in Table~\ref{tab: model compressions}. Each baseline retains the same number of nodes as the corresponding $\positivepatterns$. Although this process seems to be highly invasive, the model ablated with $\Pi^*_{+}$ achieved the highest ablated accuracy uniformly over all tasks. All baseline patterns except $\baselineconductancepatterns^e$ \footnote{We include discussions why $\baselineconductancepatterns^e$ has higher ablated accuracy than other baselines in Appendix~\ref{appendix:conductance}.} gives random guesses. For the attention-level graph $\Pi_a$, we also add a counterfactual pattern $\Pi^a_{\text{repl\_skip}}$ where we replace each skip node in $\Pi^a_+$ with its all $A$ corresponding attention heads. However, with much fewer nodes ablated, $\Pi^a_{\text{repl\_skip}}$ still produces random ablated accuracies. This corroborates Finding I in Sec.~\ref{sec: path visualization} that the skip connections relaying important information directly which attention block cannot replace. In summary, patterns refined by GPR account for the model's information flow more sufficiently compared to all baselines.

%%%%% the following part can be included in the appendix
% The relatively higher ablated accuracy of $\baselineconductancepatterns$ is also expected: pattern influence (Def. \ref{def: mppi}) using our settings of $\mathcal{D}$ and QoI, with exactly one internal node, reduces to conductance\cite{dhamdhere2018important},  therefore picking most influential node per layer using conductance is likely to achieve similar patterns. 

% However, this gap in ablated accuracy between $\baselineconductancepatterns$ with $\Pi^e_+$, albeit small, shows the utility of the GPR algorithm over a comparable layer-based approach. 

%%%%%%
% We also analyze a counterfactual pattern $\Pi^a_{repl\_skip}$ where we replace each skip node in $\Pi^a_+$ with its $|A|$ corresponding attention heads, enlarging the pattern size while also producing random ablated accuracies. This corroborates the observations in Sec.~\ref{sec: path visualization} where the skip connections play an important role in directly copying information from input to the last layer which attention block cannot replace. In summary, \textbf{influence patterns account for model performance more accurately than those derived from attention-based and layer-based methods.} 

% \subsection{Sparsity of Patterns}\label{sec:conc}
% \input{sec_eval_conc_table}
% Patterns found by GRP are sufficient enough (at least relative to baselines) to capture the BERT's behavior in the example tasks. 
%it still remains unclear how much influence actually flows through the selected pattern. 
\noindent\textbf{Sparsity of Patterns.} A good explanation should not only account for the model performance, but also be sparse relative to the entire model semantics so as to be interpretable to humans. In this section, we quantify the sparsity of extracted patterns using two metrics:  \emph{path share} and \emph{concentration}. With on average more than 10 words per sentence, the embedding and attention-level graphs contain at least $10^6$ and $10^{11}$ individual paths, respectively. % With an naive assumption that all these patterns are uniformly important then we have $C^\pi_{+} \approx$ $2^{-14}$ for a randomly picked pattern $\pi$ (when $*=e$) for example.
The totality of these paths represent the entire semantics of the BERT model. \emph{Path share}(i.e. $\text{share}(\positivepatterns^*) \defeq |\gamma_*(\Pi^*_{+})|/|\mathcal{P}^*|$), is defined as the number of paths in a pattern over the total number of paths in the entire computational graph.  \emph{concentration}, on the other hand, is defined as the proportion of negative/positive pattern influence over the total positive/negative influence. It represents the average ratio between the blue/purple bars and the orange bars in Figure~\ref{fig:intro}. % We quantify these goals and evaluate them on BERT.

Table \ref{tab: model compressions}(Right) shows that the abstracted patterns contain only a small share of paths while accounting for a large
portion of both positive and negative influence.
% That is, 
% \textbf{influence patterns cover a large portion of influence conveying a concept relative to their sparsity, making them an appropriate tool for human-facing explanations}.
In linguistic tasks, the embedding level abstracted pattern has a \textit{concentration} around $0.3$ (conc.(${\pmpatterns^e}$)), indicating that the input concept flows through single internal embeddings in each layer, instead of distributed to many words. Zooming in on the attention-level graph, a relative high concentration conc.(${\positivepatterns^a}$), suggests that between the internal embeddings of adjacent layers, influence is also more concentrated to either one attention head or the skip connection. We speculate the much lower conc.(${\positivepatterns}$) in SA is due to (1) the much larger model for the SA task with more layers and attention heads (2) sentiment information may be more diverse and complex than the information needed to encode syntax agreements, therefore input information may flow in a more distributed way. However, despite low concentration of pattern influence, the extracted patterns for SA are still effective in capturing the model performance as shown in Table~\ref{tab: model compressions}(Left).
% The difference in concentration across subtasks, however, suggests that influence flow in some tasks are more distributed than other, either across different words in certain layers, or across different attention heads/skip connections.
% \caleb{add another metric: share of skip connections nodes}

\textbf{Attention heads in $\Pi^a$ vs. Attention weight value.}
% \pxm{Rephrase as "modulation rate"?}
Another observation is that the attention head nodes in $\Pi^a$ found by GPR aligns, to greater extent than random ($1/|A|$), with the head of the largest attention weight along the corresponding embedding-level edge, as shown by \textit{alignment rate} in Table~\ref{tab: model compressions}(Left). This partial alignment is expected since the Jacobian between two nodes correlates with the coefficients (weights) in the linear attention mechanism(product rule), while the opposite is also expected since (1) attention weights themselves are not fixed model parameters thus part of the gradient flow, (2) model components other than attention blocks come between two adjacent layers (e,g. dense layer \& skip connections). In other words, attention weights are correlated, but not equivalent to gradient-based methods as explanations.

\section{Related Work}\label{sec:related}
% \paragraph{Analysis of Linguistic Knowledge in BERT} \zifan{Related work should be no longer than
% 3/4 of the current version} 
Previous work has shown the encoding of syntactic dependencies such
as SVA in RNN Language models \citep{linzen2016assessing,hupkes2018visualisation, lakretz2019emergence,jumelet2019analysing}.
% \sm{Diagnostic classifier-based analyses} 
More extensive work has since been done on
transformer-based architectures such as BERT.
% Previous work has shown the encoding of linguistic knowledge within BERT\citep{devlin2018bert}.
Probing classifiers has shown BERT encodes
many types of linguistic knowledge\citep{elazar2020bert, hewitt2019designing, tenney2019bert,
  tenney2019you, jawahar2019does, klafka2020spying,liu2019linguistic, lin2019open, coenen2019visualizing, hewitt2019structural}. \cite{goldberg2019assessing} discovers that
SVA and RA in complex clausal structures is better represented in BERT compared to an RNN model.
% This is partially explained by \cite{coenen2019visualizing, hewitt2019structural} which show that
% contextual embeddings in BERT can encode syntactic structures hierarchically similar to a dependency tree. 
% However all these analyses are done on frozen contextual
% embedding layers; the exact causal mechanism of how a concept is encoded from input to output is not explored.

\sm{Local edge-based analyses} A line of related works analyze self-attention weights of BERT\citep{clark2019does,vig2019analyzing, lin2019open, ethayarajh2021attention}, where attention
heads are found to have direct correspondences with specific dependency relations. \cite{abnar2020quantifying} and \cite{wu2020structured}, with which we compare in this paper, propose using attentions to quantify information flows in BERT. Attention weights as interpretation devices, however,  have been
controversial~\citep{serrano2019attention} and empirical analysis has
shown that attention can be perturbed or pruned while retaining performance~\citep{kovaleva2019revealing,michel2019sixteen,voita2019analyzing}. 
Our work demonstrates that attention mechanisms are only part of the computation graph, with each
attention block complemented by other model components such as dense layer and skip connections; axiomatically justified influence patterns, however, can attribute to the whole computation graph. 
The strong influence passing through skip connections also corroborates the findings of
\cite{brunner2020identifiability} which finds input tokens mostly retain their identity.
\sm{compression} Besides pruning attentions, other works\cite{prasanna2020bert, sanh2019distilbert,
  jiao2019tinybert} also show that BERT is overparametrized and can be greatly compressed. Our work
to some extent corroborates that point by pointing to the sparse gradient flow, while employing
ablation studies only to verify the sufficiency of the extracted patterns.

A closely related and concurrent work\cite{pmlr-v139-dong21a} also shows the importance of skip connections and dense(MLP) layers by decomposing and analyzing forward-pass computations in self-attention modules. Our work, in comparison, introduces a gradient-based method that can be generalized to any model as long as they are differentiable. We also focus on how influential patterns can help us understand information flow of specific NLP tasks.

 \sm{Describe ACL work.} Recent work introducing influence
paths~\citep{DBLP:conf/acl/LuMLFD20} offers another form of explanation. 
% Influence paths quantify
% the effect of model inputs on model outputs that traverse \emph{a particular} path through a model.
% The approach refines the influence-directed explanations framework of~\cite{leino2018influence}
% which instruments a neural model with input a distribution of interest (e.g. datasets meant to
% exercise a concept such as SVA) and an output quantity of interest (e.g. a measure of SVA) in order
% to compute a gradient-based measure of input influence (referred to as attribution) such as
% Integrated Gradient~\citep{sundararajan2017axiomatic} and to discover internal model elements most
% responsible (having highest influence) on the given concept. 
Lu et al.~\cite{DBLP:conf/acl/LuMLFD20}
decomposed the attribution to path-specific quantities localizing the implementation of the given
concept to paths through a model. The authors demonstrated that for LSTM models, a single path is
responsible for most of the input-output effect defining SVA. Directly applying individual paths 
to transformer-based models like BERT, however, 
results in an intractable number of paths to enumerate due to the huge number of computation edges in BERT.

% and explored the effects of unhelpful
% nouns which showed negative influence on SVA. We describe the limitations of this methodology when
% applied to BERT in~\sref{sec:methods}.

% Other interpretation work for BERT propose model-specific methods for NLP applications such as
% question answering\citep{de2020decisions} and sentiment analysis\citep{jin2019towards}. In this
% paper we focus on tasks concerning syntactic concepts justified by linguistic and grammatical
% rules, which are evaluated by the original BERT model (instead of the fine-tuned models).

% Interactions between words in negation in SA has been studied in prior work, mostly in the input domain\cite{chen2020ls, chen2020generating}. 
% Our method complementarily pinpoints where such interaction occurs inside the model. \caleb{more citations?} 
\section{Limitations and Future Work}\label{sec: limitation}
We believe there are three limitations to address in future work. (1) GPR algorithm does not guarantee absolute optimality, however, we provide empirical evidence in support of the searching algorithm in Appendix~\ref{appendix: optim-GPR}. (2) For interpretability reasons, we compute GPR by picking one node per guiding set, while more complicated information may distribute to multiple internal embeddings or attention heads. (3) The findings in Sec.~\ref{sec: path visualization} would benefit from more quantitative analysis to support more general claims(instead of speculations) on BERT's handling of various linguistic/semantic concepts. However, we will explore these limitations in future work and release our code and hope the proposed methods will serve as an insightful tool in future exploration. 
 
 \section{Conclusion}\label{sec:conclusion}
% \todo{fix and update}
We demonstrated influence patterns for explaining information flow in BERT. We highlighted the importance of skip connections and BERT's potentially mishandling of various concepts through visualized patterns. We inspected the relation between consistency of patterns across instances with model performance and quantitatively validated pattern's sufficiency in capturing model performance.
%We demonstrated BERT's contextualization in the two tasks using our methodology. Our formalism and methods are general enough to apply to the analysis of other aspects of BERT and potentially other transformer models.

\paragraph{Acknowledgement}
This work was developed with the support of NSF grant CNS-1704845. The U.S. Government is authorized
to reproduce and distribute reprints for Governmental purposes not
withstanding any copyright notation thereon. The views, opinions,
and/or findings expressed are those of the author(s) and should not
be interpreted as representing the National Science
Foundation or the U.S. Government. We gratefully acknowledge the support of NVIDIA Corporation with the donation of the Titan V GPU used for this work.
\newpage
\bibliographystyle{plain}
\bibliography{ref}

\newpage
\section*{Checklist}

% % %%% BEGIN INSTRUCTIONS %%%
% % The checklist follows the references.  Please
% % read the checklist guidelines carefully for information on how to answer these
% % questions.  For each question, change the default \answerTODO{} to \answerYes{},
% % \answerNo{}, or \answerNA{}.  You are strongly encouraged to include a {\bf
% % justification to your answer}, either by referencing the appropriate section of
% % your paper or providing a brief inline description.  For example:
% % \begin{itemize}
% %   \item Did you include the license to the code and datasets? \answerYes{See Section~\ref{gen_inst}.}
% %   \item Did you include the license to the code and datasets? \answerNo{The code and the data are proprietary.}
% %   \item Did you include the license to the code and datasets? \answerNA{}
% % \end{itemize}
% % Please do not modify the questions and only use the provided macros for your
% % answers.  Note that the Checklist section does not count towards the page
% % limit.  In your paper, please delete this instructions block and only keep the
% % Checklist section heading above along with the questions/answers below.
% % %%% END INSTRUCTIONS %%%

\begin{enumerate}

\item For all authors...
\begin{enumerate}
  \item Do the main claims made in the abstract and introduction accurately reflect the paper's contributions and scope?
    \answerYes{}
  \item Did you describe the limitations of your work?
    \answerYes{}, in Sec.~\ref{sec: limitation}
  \item Did you discuss any potential negative societal impacts of your work?
    \answerYes{}, in Appendix~\ref{appendix: impact}
  \item Have you read the ethics review guidelines and ensured that your paper conforms to them?
    \answerYes{}
\end{enumerate}

\item If you are including theoretical results...
\begin{enumerate}
  \item Did you state the full set of assumptions of all theoretical results?
    \answerYes{} Only one theoretical result (Prep. 1 in in \sref{sec:methods})
	\item Did you include complete proofs of all theoretical results?
    \answerYes{} A simple proof is included in Appendix~\ref{appendix: distributional influence}
\end{enumerate}

\item If you ran experiments...
\begin{enumerate}
  \item Did you include the code, data, and instructions needed to reproduce the main experimental results (either in the supplemental material or as a URL)?
    \answerYes{} In \sref{sec:evaluation}. More details in the appendix~\ref{appendix: exp details}.
  \item Did you specify all the training details (e.g., data splits, hyperparameters, how they were chosen)?
    \answerYes{}In \sref{sec:evaluation}. More details in the appendix ~\ref{appendix: exp details}.
	\item Did you report error bars (e.g., with respect to the random seed after running experiments multiple times)?
    \answerYes{}. The only random experiments are done for Table~\ref{tab: model compressions}, the error statistics are included in \ref{appendix:random}.
    \item Did you include the total amount of compute and the type of resources used (e.g., type of GPUs, internal cluster, or cloud provider)?
    \answerYes In \sref{sec:evaluation}. More details in the appendix~\ref{appendix: exp details}.
\end{enumerate}

\item If you are using existing assets (e.g., code, data, models) or curating/releasing new assets...
\begin{enumerate}
  \item If your work uses existing assets, did you cite the creators?
    \answerYes{} In In \sref{sec:evaluation}. 
  \item Did you mention the license of the assets?
    \answerYes{}. in the appendix~\ref{appendix: exp details}.
  \item Did you include any new assets either in the supplemental material or as a URL?
    \answerYes{}. Code included in the supplementary materials. 
  \item Did you discuss whether and how consent was obtained from people whose data you're using/curating?
    \answerNA{}.
  \item Did you discuss whether the data you are using/curating contains personally identifiable information or offensive content?
    \answerNA{}
\end{enumerate}

\item If you used crowdsourcing or conducted research with human subjects...
\begin{enumerate}
  \item Did you include the full text of instructions given to participants and screenshots, if applicable?
    \answerNA{}
  \item Did you describe any potential participant risks, with links to Institutional Review Board (IRB) approvals, if applicable?
    \answerNA{}
  \item Did you include the estimated hourly wage paid to participants and the total amount spent on participant compensation?
    \answerNA{}
\end{enumerate}

\end{enumerate}

%%%%%%%%%%%%%%%%%%%%%%%%%%%%%%%%%%%%%%%%%%%%%%%%%%%%%%%%%%%%

\newpage
\appendix

\section{Appendix: Proof of Proposition 1}
\label{appendix: distributional influence}
\noindent\textbf{Proposition 1 (Chain Rule)}\textit{
 $\mathcal{I}(\mathbf{x}, \pi) = \mathbb{E}_{\mathbf{z} \sim \mathcal{D}(\xvec)}
    \prod_{i=1}^{-1}\frac{\partial \pi_i(\mathbf{z})}{\partial \pi_{i-1}(\mathbf{z})}$ for any distribution $\mathcal{D}(\xvec)$.
}

We first show that for a pattern $\pi=[\pi_1, \pi_1, ..., \pi_k, ..., \pi_n]$ the following equation holds:
\begin{align}
    \frac{\partial \pi_n}{\partial \pi_1} = \frac{\partial \pi_n}{\partial \pi_k}\frac{\partial \pi_k}{\partial \pi_1}
\end{align}
\textit{Proof:}
Let $\gamma(\pi_1 \rightarrow \pi_n)$ be a set of paths that the pattern $\pi$ abstracts. Therefore, we have 
\begin{align}
    \frac{\partial \pi_n}{\partial \pi_1} = \sum_{p \in \gamma(\pi_1 \rightarrow \pi_n)} \prod^{-1}_{i=1} \frac{\partial p_i}{\partial p_{i-1}}
\end{align} Similarly, we have 
\begin{align}
    \frac{\partial \pi_k}{\partial \pi_1} = \sum_{p \in \gamma(\pi_1 \rightarrow \pi_k)} \prod^{-1}_{i=1} \frac{\partial p_i}{\partial p_{i-1}},\quad 
    \frac{\partial \pi_n}{\partial \pi_k} = \sum_{p \in \gamma(\pi_k \rightarrow \pi_n)} \prod^{-1}_{i=1} \frac{\partial p_i}{\partial p_{i-1}}
\end{align} Suppose $|\gamma(\pi_1 \rightarrow \pi_k)| = N_1$, $|\gamma(\pi_k \rightarrow \pi_n)| = N_2$ and we denote $\prod^{-1}_{i=1} \frac{\partial p^{(j)}_i}{\partial p_{i-1}}$ as path gradient flowing from the path $j$. Therefore,

\begin{align}
    \frac{\partial \pi_n}{\partial \pi_1} &= \sum_{p \in \gamma(\pi_1 \rightarrow \pi_n)} \prod^{-1}_{i=1} \frac{\partial p_i}{\partial p_{i-1}}\\
    &= \prod^{-1}_{i=k} \frac{\partial p^{(1)}_i}{\partial p_{i-1}} \cdot \prod^{k}_{i=1} \frac{\partial p^{(1)}_i}{\partial p_{i-1}} + \prod^{-1}_{i=k} \frac{\partial p^{(1)}_i}{\partial p_{i-1}} \cdot \prod^{k}_{i=1} \frac{\partial p^{(2)}_i}{\partial p_{i-1}} + ... + \prod^{-1}_{i=k} \frac{\partial p^{(1)}_i}{\partial p_{i-1}} \cdot \prod^{k}_{i=1} \frac{\partial p^{(N_1)}_i}{\partial p_{i-1}}\\
    &+ \prod^{-1}_{i=k} \frac{\partial p^{(2)}_i}{\partial p_{i-1}} \cdot \prod^{k}_{i=1} \frac{\partial p^{(1)}_i}{\partial p_{i-1}} + \prod^{-1}_{i=k} \frac{\partial p^{(2)}_i}{\partial p_{i-1}} \cdot \prod^{k}_{i=1} \frac{\partial p^{(2)}_i}{\partial p_{i-1}} + ... + \prod^{-1}_{i=k} \frac{\partial p^{(2)}_i}{\partial p_{i-1}} \cdot \prod^{k}_{i=1} \frac{\partial p^{(N_1)}_i}{\partial p_{i-1}}\\
    &+ ...\\
    &+ \prod^{-1}_{i=k} \frac{\partial p^{(N_2)}_i}{\partial p_{i-1}} \cdot \prod^{k}_{i=1} \frac{\partial p^{(1)}_i}{\partial p_{i-1}} + \prod^{-1}_{i=k} \frac{\partial p^{(N_2)}_i}{\partial p_{i-1}} \cdot \prod^{k}_{i=1} \frac{\partial p^{(2)}_i}{\partial p_{i-1}} + ... + \prod^{-1}_{i=k} \frac{\partial p^{(N_2)}_i}{\partial p_{i-1}} \cdot \prod^{k}_{i=1} \frac{\partial p^{(N_1)}_i}{\partial p_{i-1}}\\
    &=\sum^{N_2}_j (\prod^{-1}_{i=k} \frac{\partial p^{(j)}_i}{\partial p_{i-1}} \cdot \sum^{N_1}_{m} \prod^{k}_{i=1} \frac{\partial p^{(m)}_i}{\partial p_{i-1}})\\
    &=\sum^{N_2}_j (\prod^{-1}_{i=k} \frac{\partial p^{(j)}_i}{\partial p_{i-1}}) \cdot \sum^{N_1}_{m}( \prod^{k}_{i=1}\frac{\partial p^{(m)}_i}{\partial p_{i-1}})\\
    &=\sum_{p \in \gamma(\pi_1 \rightarrow \pi_k)} \prod^{-1}_{i=1} \frac{\partial p_i}{\partial p_{i-1}} \cdot \sum_{p \in \gamma(\pi_k \rightarrow \pi_n)} \prod^{-1}_{i=1} \frac{\partial p_i}{\partial p_{i-1}}\\
    &=\frac{\partial \pi_n}{\partial \pi_k}\frac{\partial \pi_k}{\partial \pi_1}
\end{align}

Now we prove Proposition 1.
\begin{align}
    \mathcal{I}(\mathbf{x}, \pi) &= \sum_{p \in \gamma(\pi)} \mathbb{E}_{\mathbf{z} \sim \mathcal{D}(\xvec)} \prod^{-1}_{i=1} \frac{\partial p_{i}(\mathbf{z})}{\partial p_{i-1}(\mathbf{z})}\\
    &=\mathbb{E}_{\mathbf{z} \sim \mathcal{D}(\xvec)} \sum_{p \in \gamma(\pi)} \prod^{-1}_{i=1} \frac{\partial p_{i}(\mathbf{z})}{\partial p_{i-1}(\mathbf{z})}\\
    &=\mathbb{E}_{\mathbf{z} \sim \mathcal{D}(\xvec)}
    \prod_{i=1}^{-1}\frac{\partial \pi_i(\mathbf{z})}{\partial \pi_{i-1}(\mathbf{z})}
\end{align}

\section{Appendix: Guided Pattern Refinement}

\subsection{Algorithms for GPR}
The pseudo code of the GPR algorithms are presented in algorithm~\ref{algo-gpre} and \ref{algo-gpra}. 
\label{appendix: pseudo-code}
\begin{algorithm}[t]
	\SetAlgoLined
	\KwResult{Significant Path $\pi^e$}
	initialization\;
	$\mathbf{x} \sim \text{Input Tokens}$, $f \gets \text{BERT}$ \;
	$\mathcal{G}_e \gets \text{GetEmbeddingGraph}(f)$,
	$L\gets \text{GetNumberOfLayers}(f)$\;
	$N \gets\text{GetNumberOfTokens(f)}$, $m \gets \text{GetIndex}(\texttt{[MASK]})$, \;
	$n_q \gets \text{GetQoINode($m, \mathcal{G}_e$)}$, $n_w \gets \text{GetClfNode($\mathcal{G}_e$)}$\;
	$\pi^e \gets\text{OrderedSet()}$,  $\mathcal{C} \gets \{\empty\}$ \;
	$j \gets \text{GetStartingIndex()}$\tcp*{The word we start the search with}
	$n^0_j \gets \text{GetNode}(\mathcal{G}_w, \mathbf{h}^0_j)$ \tcp*{Find the corresponding node}
	$\pi^e \gets \text{Append}(\pi^e, n^0_j)$\;
	\For{$l\in\{1, ..., L-2\}$}{
		$\mathcal{C} \gets \{\empty\}$\;
		\For{$i \in \{0, ..., N-1\}$}{
			$n^l_i \gets \text{GetNode}(\mathcal{G}_w, \mathbf{h}^l_i)$ \;
			$\pi_t \gets \text{Append}(\pi, n^l_i)$\;
			$\mathcal{C} \gets \mathcal{C} \cup \{\pi_t\}$ \;
		}
		$\pi^e \gets \arg\max_{\pi' \in \mathcal{C}} \mathcal{I}(\mathbf{x}|\pi', \pi'_{-1} \rar^{\mathcal{P}} n_q )$
	}
	$n^L_m \gets \text{GetNode}(\mathcal{G}_w, \mathbf{h}^{L-1}_m)$\;
	$\pi^e \gets \text{Append}(\pi^e, n^L_m)$\;
	$\pi^e \gets \text{Append}(\pi^e, n_w)$\;
	$\pi^e \gets \text{Append}(\pi^e, n_q)$\;
	\caption{Guided Pattern Refinement in the Embedding-level Graph (GPR-e)}
	\label{algo-gpre}
\end{algorithm}

\begin{algorithm}[ht]
	\SetAlgoLined
	\KwResult{Significant Path $\pi^a$}
	initialization\;
	$\mathbf{x} \sim \text{Input Tokens}$, $f \gets \text{BERT}$ \;
	$\mathcal{G}_e \gets \text{GetEmbeddingGraph}(f)$,
	$\mathcal{G}_a \gets \text{GetAttentionGraph}(f)$\;
	$m\gets\text{GetIndex}(\texttt{[MASK]})$\;
	$n_q \gets \text{GetQoINode($m, \mathcal{G}_e$)}$\;
	$L\gets \text{GetNumberOfLayers}(f)$, $N \gets\text{GetNumberOfTokens(f)}$ \;
	$\pi^e \gets \text{GPR-e}(\mathbf{x}, \mathcal{G}_e)$\tcp*{Find embedding-level path first}
	$\mathcal{C} \gets \{\empty\}$\;
	\For{$i\in\{0, ..., |\pi^e|-2\}$}{
  \eIf{$\text{ExistAttentionBlock}$($\pi^e_i, \pi^e_{i+1}$)\tcp*{Check if this is a Transformer Layer}}{
    $\mathcal{A}_i \gets \text{GetHeadsBetween}(\mathcal{G}_a, \pi^e_i, \pi^e_{i+1}) $\tcp*{Get all attention heads and the skip connetion node}
    $\pi^e_{head} \gets \text{Slice}(\pi^e_0, \pi^e_i)$\tcp*{Take a slice between two nodes }
    $(a^*_i, c^*_i) \gets \arg\max_{(a_i, c_i) \in \mathcal{A}_i}\mathcal{I}(\mathbf{x}|\pi^e_{head}\cup\{a_i, c_i\}, c_i \rar^{\mathcal{P}} n_q )$\;
    $\mathcal{C} \gets \mathcal{C} \cup \{(a^*_i, c^*_i) \}$\;
  }{
  continue\;
  }
	}
		$\pi^a \gets \text{InsertNode}(\pi^e, \mathcal{C})$\tcp*{Insert attention nodes into the embedding-level path at the corresponding place}
	\caption{Guided Pattern Refinement in the Attention-level Graph (GPR-a)}
	\label{algo-gpra}
\end{algorithm}
\subsection{Optimality of GPR}
\label{appendix: optim-GPR}
% \todo{some arguments that a optimal searching algorithm using dynamic searching is not possible unless it's done on all dimensions}. 
% \todo{statistic test on the optimality of the algorithm}
 The definition of pattern influence does not allow for polynomial-time searching algorithms such as dynamic programming. (Such an algorithm is possible for simple gradients/saliency maps but not for integrated gradients due to the expectation sum over the multiplication of Jacobians along all edges). As for the optimality of the polynomial-time greedy algorithm, we hereby include a statistical analysis by randomly sampling 1000 alternative patterns for 100 word patterns in the \textit{SVA-obj} task and sentiment analysis task (SST2). The pattern influence of those random paths shows that the patterns extracted GPR are (1) more influential than 999.96 and 1000(all) random patterns, averaged across all 100 word patterns evaluated for embedding-level attention-level patterns, respectively for \textit{SVA-obj}; the same holds for sentiment analysis (1000 and 1000); (2) The extracted pattern influences are statistically significant assuming all randomly sampled patterns' influences follow a normal distribution, with $p=$2e-10 for and $p = 0$ for embedding-level and attention-level patterns, respectively for \textit{SVA-obj}; the same holds for sentiment analysis ($p=0$ for and $p = 0$). In other words, the pattern influence values are far more significant than a random pattern, also confirmed by the high concentration values shown in ~\sref{sec: gsig}. 
 
\section{Appendix: Baseline Patterns}
\subsection{Baseline: Attention-based baseline}
\label{appendix: Attention Path}
We introduce the implementation details of attention-based patterns, inspired by \cite{abnar2020quantifying} and \cite{wu2020structured}. Consider a BERT model with $L$ layers, between adjacent layers $l$ and $l+1$, the attention matrix is denoted by $M_l \in \mathbb{R}^{A\times N \times N}$ where $A$ is the number of attention heads and $K$ is the number of embeddings. Each element of $M_l[a,i,j]$ is the attentions scores between the $i$-th embedding at layer $l$ and the $j$-th embedding at layer $l+1$ of the $a$-th head such that $\sum_i M_l[a, i, j] = 1$. We average the attentions scores over all heads to lower the dimension of $M$: $\Tilde{M}_l \defeq \frac{1}{A}\sum_a M_l[a]$. 
% A number of recent work~\cite{} treat the attention score between two embeddings proportional to how strong the embedding from layer $l$ is going to contextualize with the the embedding in layer $l+1$, therefore, 
We define the baseline attention path as a path where the product of each edge in this path is the maximum possible score among all paths from a given source to the target.
\begin{definition}[Attention-based Pattern]
\label{def: attention path}
Given a set of attention matrices $\Tilde{M}_0, \Tilde{M}_1, ..., \Tilde{M}_{L-1}$, a source embedding $x$ and the quantity of interest node $q$, an attention-based pattern $\baselineattentionpatterns$ is defined as
\begin{align*}
    \baselineattentionpatterns \defeq
    \{x, h^1_*, h^2_*, ..., h^{L-2}_*, q\}
\end{align*} where 
\begin{align*}
    & h^1_*, h^2_*, ..., h^{L-2}_* \\
    & = \arg\max_{j_1, j_2, ..., j_{L-2}} P(h^1_{j_1}, h^2_{j_2}, ..., h^{L-2}_{j_{L-2}})\\
    &P = \Tilde{M}_0[s,j_1]\Tilde{M}_{L-1}[j_{L-2},t]\prod^{L-2}_{l=1} \Tilde{M}_0[j_{l},j_{l+1}] 
\end{align*}
\end{definition}

\cite{abnar2020quantifying} considers an alternative choice $\hat{M}_l \defeq 0.5I + 0.5\Tilde{M}_l$ to model the skip connection in the attention block where $I$ is an identity matrix to represent the identity transformation in the skip connection, which we use GPR in Sec.~\ref{sec:evaluation}, we use $\hat{M}_l$ to replace $\Tilde{M}_l$ when returning the attention-based pattern. Dynamic programming can be applied to find the maximum of the product of attentions scores and back-trace the optimal nodes at each layer to return $h^1_*, h^2_*, ..., h^{L-2}_*$. 
% In the Sec.~\ref{sec:evaluation}, the source $s$ is an input embedding that is not \texttt{[MASK]} and the target embedding at the final layer is always \texttt{[MASK]}. 
\subsection{Baseline: Conductance and Distributional Influence}\label{appendix:conductance}
% \todo{Zifan: definitions of conductance and distributional influence and how pattern is extracted}

We find patterns consisting of nodes that maximizes \emph{condutance}~\cite{dhamdhere2018important} and \emph{internal influence}~\cite{leino2018influence} to build baseline methods $\baselineconductancepatterns$ and $\baselineinfluencepatterns$, respectively. Definitions of these two measurements are shown as follows:

\begin{definition}[Conductance~\cite{dhamdhere2018important}]
Given a model $f: \mathbb{R}^d \rightarrow \mathbb{R}^n$, an input $\mathbf{x}$, a baseline input $\mathbf{x}_b$ and a QoI $q$, the conductance on the output $\mathbf{h}$ of a hidden neuron is defined as
\begin{align}
    c_q(\mathbf{x}, \mathbf{x}_b, \mathbf{h}) = (\mathbf{x} - \mathbf{x}_b)\circ \sum_i \mathbf{1}_i \circ \mathbb{E}_{\mathbf{z} \sim \mathcal{U}}[ \frac{\partial q(f(\mathbf{z}))}{\partial \mathbf{h}(\mathbf{z})}\frac{\partial \mathbf{h}(\mathbf{z})}{\partial z_i}]
\end{align} where $\mathbf{1}_i$ is a vector of the same shape with $\mathbf{x}$ but all elements are 0 except the $i$-th element is filled with 1. $\mathcal{U} := \textit{Uniform}(\overline{\mathbf{x}_b \mathbf{x}})$.
\end{definition}

\begin{definition}[Internal Influence~\cite{leino2018influence}]
Given a model $f: \mathbb{R}^d \rightarrow \mathbb{R}^n$, an input $\mathbf{x}$, a QoI $q$ and a distribution of interest $\mathcal{D}$, the internal influence on the output $\mathbf{h}$ of a hidden neuron is defined as
\begin{align}
    \chi_q(\mathbf{x}, \mathbf{h}) =  \mathbb{E}_{\mathbf{z} \sim \mathcal{D}}[ \frac{\partial q(f(\mathbf{z}))}{\partial \mathbf{h}(\mathbf{z})}]
\end{align}
\end{definition}

In this paper, we use a uniform distribution over a path $c = \{\xvec + \alpha(\xvec-\xvec_b), \alpha \in [0, 1]\}$ from a user-defined baseline input $\xvec_b$ (\texttt{[MASK]}) to the target input $\xvec$, reducing the internal influence to Integrated Gradient(IG)~\citep{sundararajan2017axiomatic} if we multiply the $(\xvec-\xvec_b)$ with the distributional influence, as mentioned in Sec.~\ref{sec:evaluation}. IG is an extention of Aumann Shapley values in the deep neural networks, which satisfies a lot of natural axioms: efficiency, dummy, path-symmetry, etc.\cite{sundararajan2020shapley}. Choices of DoIs, besides the one used in IG, include Gaussian distributions with mean $\xvec$~\citep{smilkov2017smoothgrad} and Uniform distribution around $\xvec$~\cite{wang2020smoothed} are shown to have other nice properties such as robustness against adversarial perturbations.

The relatively higher ablated accuracy of $\baselineconductancepatterns$ is expected: pattern influence (Def. \ref{def: mppi}) using our settings of $\mathcal{D}$ and QoI, with exactly one internal node, reduces to conductance\cite{dhamdhere2018important},  therefore picking most influential node per layer using conductance is likely to achieve similar patterns. However, this gap in ablated accuracy between $\baselineconductancepatterns$ with $\Pi^e_+$, albeit small, shows the utility of the GPR algorithm over a comparable layer-based approach. 

\subsection{Baseline: Variation Statistics for random patterns}\label{appendix:random}

\begin{table}[h]
\centering
\resizebox{0.7\linewidth}{!}{%
    \centering
    \begin{tabular}{c|cccc|cc|c}
    \hline
     \multirow{2}{*}{Patterns} & \multicolumn{4}{c|}{SVA}& \multicolumn{2}{c|}{RA} & \multirow{2}{*}{SA}\\
     \cline{2-7}
     & Obj. & Subj. & WSC & APP &  NA & GA \\
      \hline
     $\Pi^e_{+}$(ours)&\textbf{.96} &\textbf{.99} &\textbf{1.0} &\textbf{.96} &\textbf{.98} &\textbf{.99} & \textbf{.97}\\
     $\Pi^e_{\text{rand}}$ &.55$\pm.017$ &.60$\pm.014$&.55$\pm.016$&.55$\pm.016$ &.55$\pm.015$ &.56$\pm.014$ & .52$\pm.030$\\
     $\Pi^e_{\text{attn}}$ &.61&.55 &.49 &.63 &.56 &.65 &  .47\\
     $\Pi^e_{\text{cond}}$ &.93&.91 &.88 &.90 &.71 &.95 &  .94\\
     $\Pi^e_{\text{inf} }$ &.66&.71 &.50 &.56 &.68 &.50 &  .49\\

    \hline
     $\Pi^a_{+}$(ours) &\textbf{.70} &\textbf{.62} &\textbf{.56} &\textbf{.65} &\textbf{.78} &\textbf{.89} & \textbf{.86}\\
     $\Pi^a_{\text{rand}}$ &.50$\pm.010$ &.50$\pm.008$ &.50$\pm.009$ &.50$\pm.011$ &.50$\pm.011$ &.50$\pm.008$ & .49$\pm.022$\\
     \hline
     $\Pi^a_{\text{repl\_skip}}$ &.50 &.58 &.54 &.52 &.55 &.50 & .48\\
    %  acc.($\pi^a_{baseline,i}$) & & & & & & \\
    \hline
    original &.96 &1.0 &1.0 &1.0 &.83 &.73 & .92\\
    \hline
    \end{tabular}}

\caption{ Ablated accuracies(Table \ref{tab: model compressions} Left) of $\positivepatterns$ and baseline patterns expanded with standard deviations for the random baseline patterns over 50 runs.}
\label{tab: variation}
% \vspace{-6mm}
\end{table}

% \newpage
% comment out the appendix for now cuz it's too slow. 

% \section{Appendix: Computational Graphs}

\section{Appendix: Experiment Details}
\label{appendix: exp details}
\subsection{Task and Data details}
For linguistics tasks, example sentences and accuracy can be found in Table~\ref{tab: model_data}. We also include the accuracy of a larger BERT model ($\text{BERT}_{\text{BASE}}$) which are comparable in performance with the smaller BERT model used in this work. First 5 tasks are sampled from \cite{marvin2018targeted}(MIT license), the last task is sampled from dataset in \cite{lin2019open}(Apache-2.0 License), all datasets are constructed as an MLM task according to \cite{goldberg2019assessing}(Apache-2.0 License). In order to compute quantitative results, we sample sentences with a fixed length from each subtask. SA data (SST-2) are licensed under GNU General Public License. The BERT pretrained models\cite{devlin2018bert} are licensed under Apache-2.0 License.
\begin{table}[t]
	\centering
    \resizebox{0.8\linewidth}{!}{%
\begin{tabular}{p{2cm}|p{0.7cm}|p{5.8cm}|p{1cm}|p{1cm}}
% \begin{tabular}{l|l|l|l|l}

Task & Type &  Example & \makecell{BERT \\Small} & \makecell{BERT \\Base}   \\
\hline
\hline
SVA & & & &  \\
\hline
\multirow{4}{0in}{Object Relative Clause} & SS  & \multirow{4}{6cm}{\centering the author that the guard likes [MASK(is/are)] young} & 1& 1\\
 & SP  &  & 0.92 & 0.96\\
 & PS  &  & 0.9 & 0.98\\
 & PP  &  & 1 & 1\\
 \hline
\multirow{4}{0in}{Subject Relative Clause} & SS  & \multirow{4}{6cm}{\centering the author that likes the guard [MASK(is/are)] young} & 1&  1\\
 & SP  &  &  1 & 0.96\\
 & PS  &  &  1 & 0.98\\
 & PP  &  &  1 & 1\\
 \hline
\multirow{4}{0in}{ Within Sentence Complement} & SS  & \multirow{4}{6cm}{\centering the mechanic said the author [MASK(is/are)] young} & 1& 1\\
 & SP  &  &  1& 1\\
 & PS  &  &  1& 1\\
 & PP  &  &  1& 1\\
  \hline
\multirow{4}{0in}{Across Prepositional Phrase} & SS  & \multirow{4}{6cm}{\centering the author next to the guard [MASK(is/are)] young} & 1& 0.99\\
 & SP  &  &  1& 0.98\\
 & PS  &  &  0.98& 0.98 \\
 & PP  &  &  1& 1 \\
 \hline
 \hline
Reflexive Anaphora & & & & \\
\hline
\multirow{4}{0in}{Number \\Agreement} & SS  & \multirow{4}{6cm}{\centering the author that the guard likes hurt [MASK(himself/themselves)]}  & 0.66& 0.6\\
 & SP  &  &  0.66& 0.74\\
 & PS  &  & 0.83 & 0.83\\
 & PP  &  & 1 & 0.96\\
 \hline
\multirow{4}{0in}{Gender \\Agreement} & MM  & \multirow{4}{6cm}{\centering some wizard who can dress our man can clean [MASK(himself/herself)]} & 0.78 & 1\\
 & MF  &  &  0.32& 0.96\\
 & FF  &  &  1& 0.9\\
 & FM  &  &  0.8& 0.66\\
 \hline
\end{tabular}}
	\caption{Example of each agreement task and their performance on two BERT models.}
	\label{tab: model_data}
\end{table}

\label{appendix: Linguistic tasks}
\label{appendix: performance}
% \paragraph{Sentiment Analysis} The reason we only sample shorter sentences from SST2 is that

\subsection{Experiment Setup}
\label{appendix: setup}
Our experiments ran on one Titan V GPU with tensorflow\cite{tensorflow2015-whitepaper}. On average GPR for attention-level and embedding-level take around half a minute to run for each instance in SVA and RA, and around 1 min for SA, with 50 batched samples to approximate influence. The whole quantitative experiment across all tasks takes around 1 days cumulatively. 
% \subsection{Model Compression Algorithm}
\subsection{Code submission}
We will release the code publicly once the paper is published. 
% \subsection{Convergence Check}\label{appendix:experiment-convergence}
% When the DoI of $g_q(\xvec)$ is a uniform distribution on a linear path from a baseline input $\xvec_b$ to the target input $\xvec$, \emph{completeness}~\citep{sundararajan2017axiomatic} axiom dictates that $q(f(\xvec)) - q(f(\xvec_b)) = \sum_i x_i g_q(\xvec)_i$, where $q$ is the selected QoI. However, when summation is used to approximate the expectation in practice, the RHS of the axiom does not easily converges to the LHS. In Fig.~\ref{fig: convergence}, we plot the  percentage of difference $[q(f(\xvec)) - q(f(\xvec_b)) - \sum_i x_i g_q(\xvec)_i] / (q(f(\xvec)) - q(f(\xvec_b)))$ against \textit{resolution}, the number of samples drawn from the distribution in the summation. we find the maximum number of batched samples to be 50. Therefore, $\text{BERT}_{\text{SMALL}}$ has lower approximation error compared to $\text{BERT}_{\text{BASE}}$. The much harder approximation of the larger BERT model is likely due to its complicated decision boundaries, making the outputs sensitive to small perturbations.
% \begin{figure}[h] 
%     \centering
% 	\includegraphics[width=.4\textwidth]{images/newplot.png}
% 	\caption{Resolution vs. Percentage of difference to Actual QoI between $\text{BERT}_{\text{SMALL}}$(used in this work) \& $\text{BERT}_{\text{BASE}}$ from~\cite{devlin2018bert}}
% 	\label{fig: convergence}
% \end{figure}

\section{Appendix: More Visualizations of Patterns}
\label{appendix: more viz}
% \subsection{Baseline Pattern Entropy vs. Average QoI}
% Figure~\ref{fig:baseline entropy} shows the relation discussed in Sec.~\ref{sec:entropy} for the baseline attention-based patterns. We can see that the linear relation is much weaker than the one in Sec.~\ref{sec:entropy}. 
% \label{appendix: baseline entropy}

% \begin{figure}[h]
%     \centering
%     \includegraphics[width = .35\textwidth]{images/fig3_ent_baseline.pdf}
%     \caption{Baseline Pattern Entropy vs. Average QoI for subtasks. (The entropy is overall smaller for the baseline patterns, since the patterns turn out to be consistent but not meaningful. )}
%     \label{fig:baseline entropy}
% \end{figure}

\subsection{Example Visualizations}
\label{appendix: example graphs}
\begin{figure}[ht]
\centering
  \begin{subfigure}[]{.33\textwidth}
    \centering
    \includegraphics[width=\linewidth]{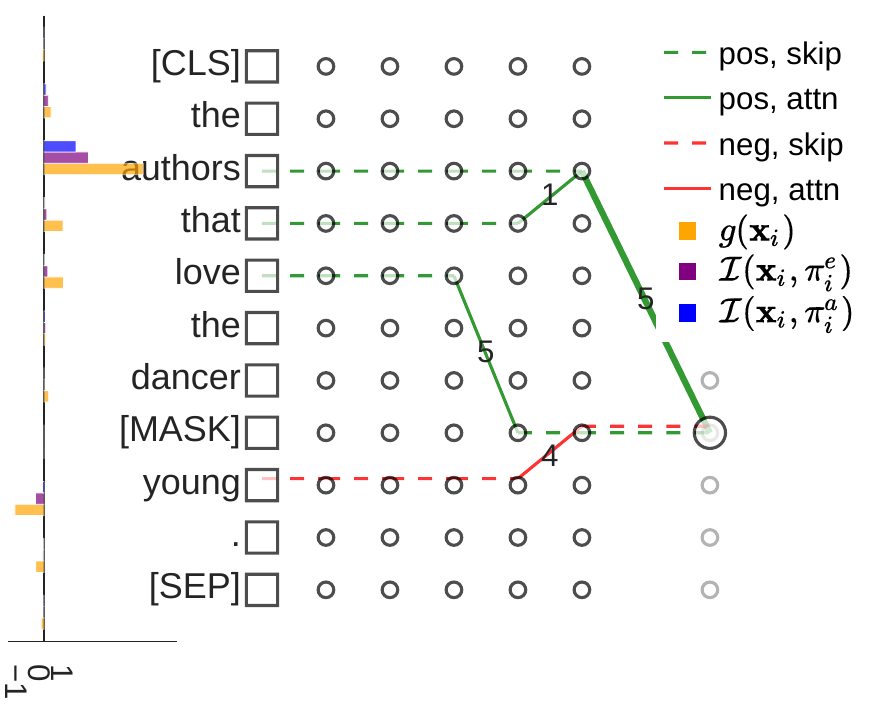}
    \caption{SVA-Subj, PS case}
  \end{subfigure}

  \begin{subfigure}[]{.33 \textwidth}
    \centering
    \includegraphics[width=\linewidth]{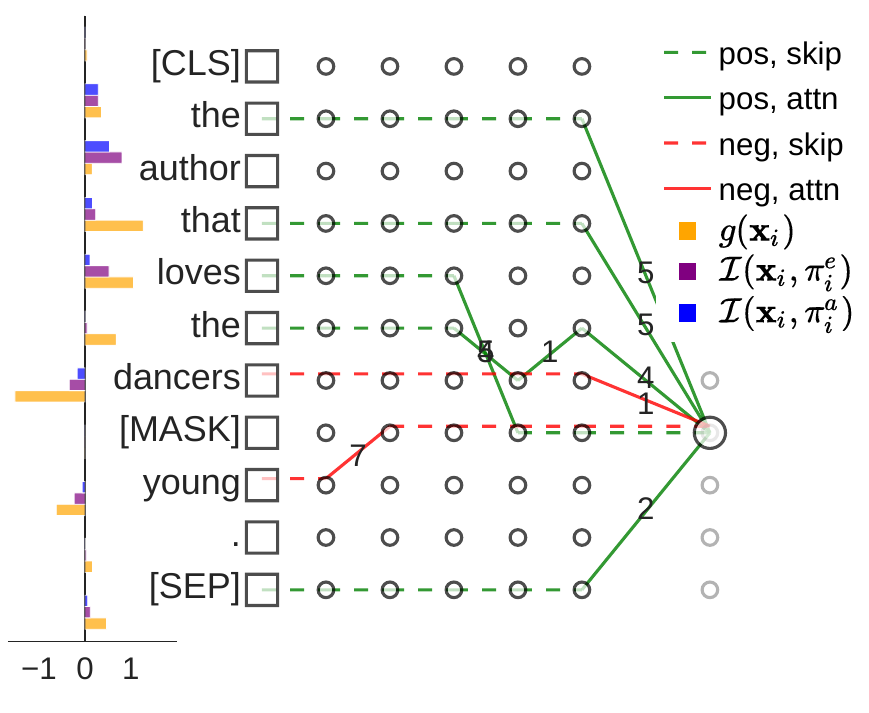}
    \caption{SVA-Subj, SP case}
  \end{subfigure}
   \caption{Examples of SVA-Subj}
  \label{fig:e1}
\end{figure}

\begin{figure}[h]
\centering
  \begin{subfigure}[]{.33\textwidth}
    \centering
    \includegraphics[width=\linewidth]{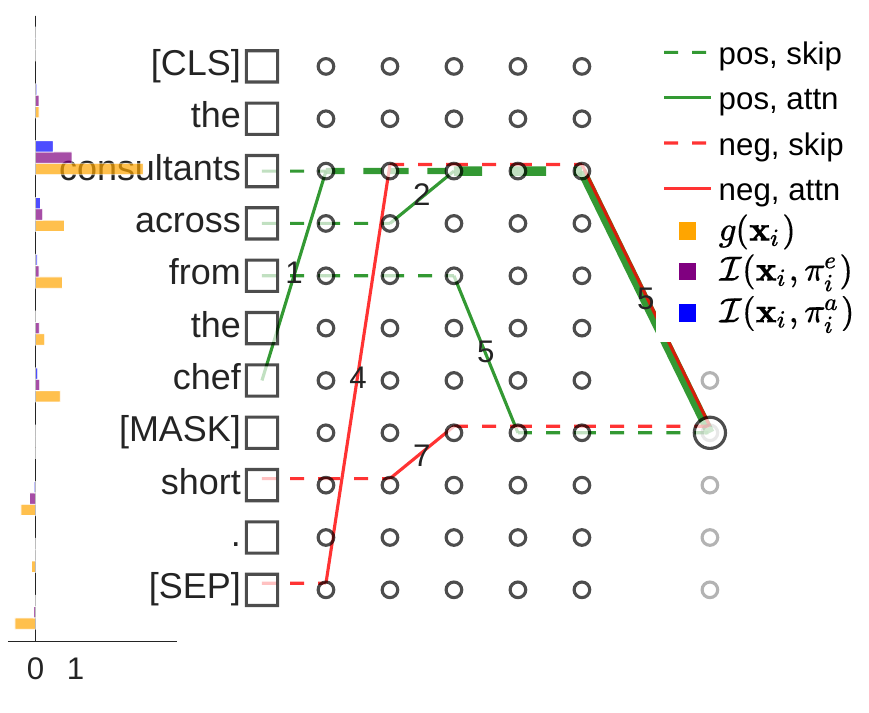}
    \caption{SVA-APP, PS case}
  \end{subfigure}
  \begin{subfigure}[]{.33 \textwidth}
    \centering
    \includegraphics[width=\linewidth]{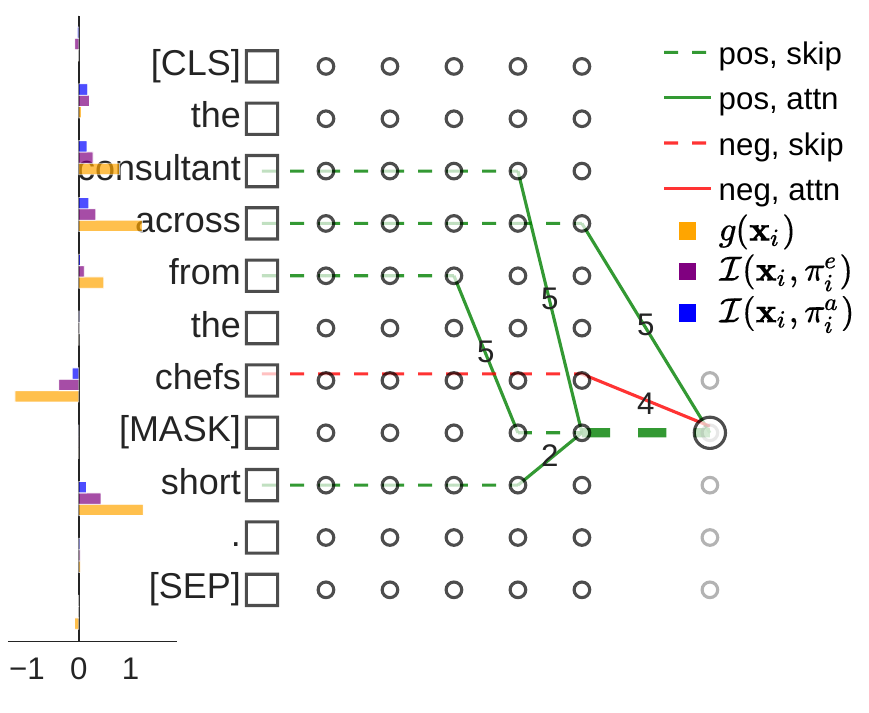}
    \caption{SVA-APP, SP case}
  \end{subfigure}
   \caption{Examples of SVA-APP}
  \label{fig:e2}
\end{figure}

\begin{figure}[h]
\centering
  \begin{subfigure}[]{.33\textwidth}
    \centering
    \includegraphics[width=\linewidth]{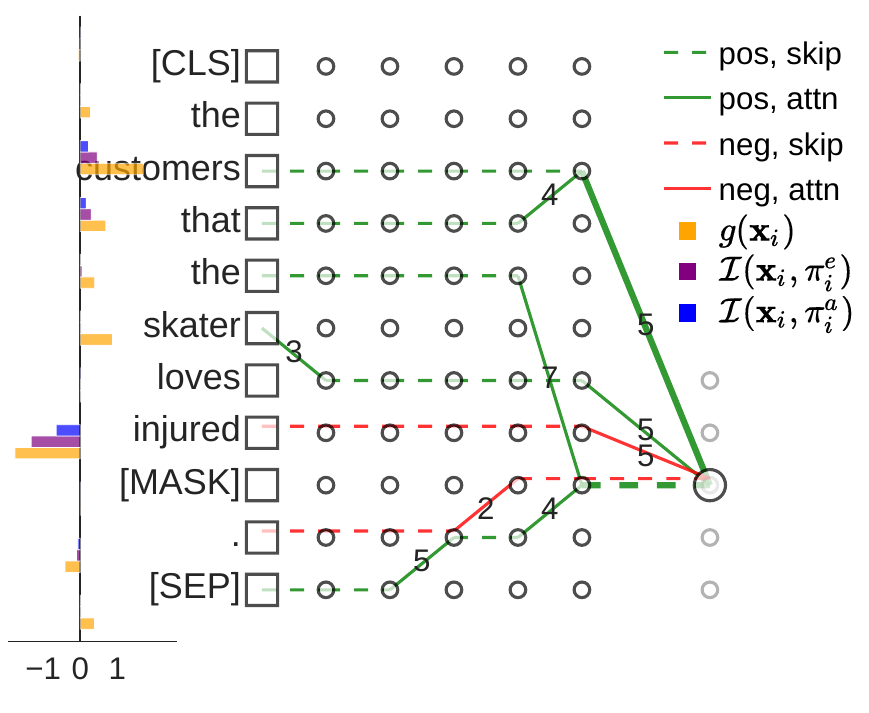}
    \caption{RA-NA, PS case}
  \end{subfigure}
  \begin{subfigure}[]{.33 \textwidth}
    \centering
    \includegraphics[width=\linewidth]{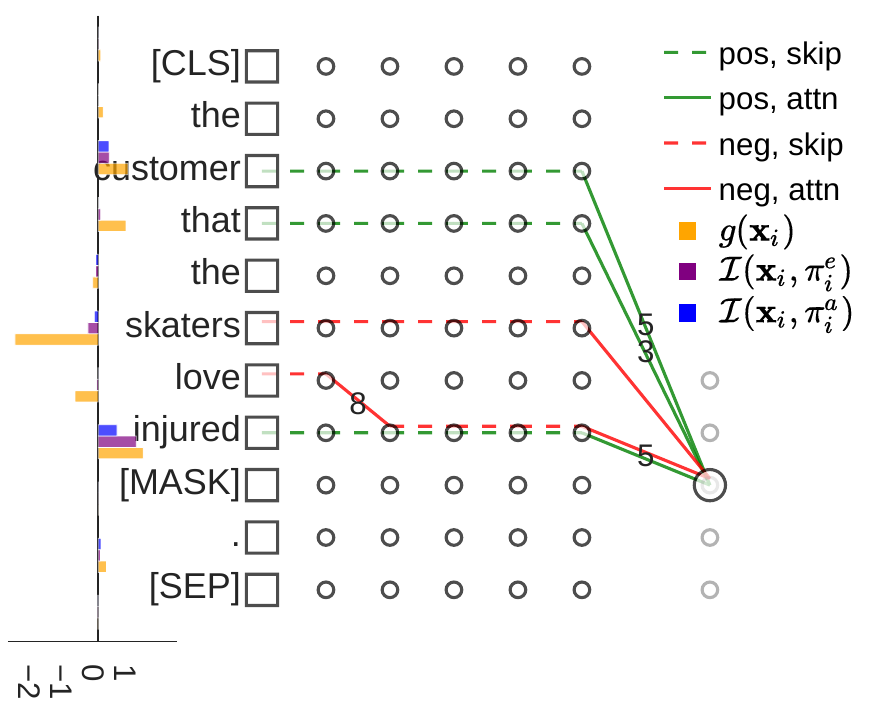}
    \caption{RA-NA, SP case}
  \end{subfigure}
  \caption{Examples of RA-NA}
  \label{fig:e3}
\end{figure}

\begin{figure}[h]
	\centering
	\includegraphics[width=\linewidth]{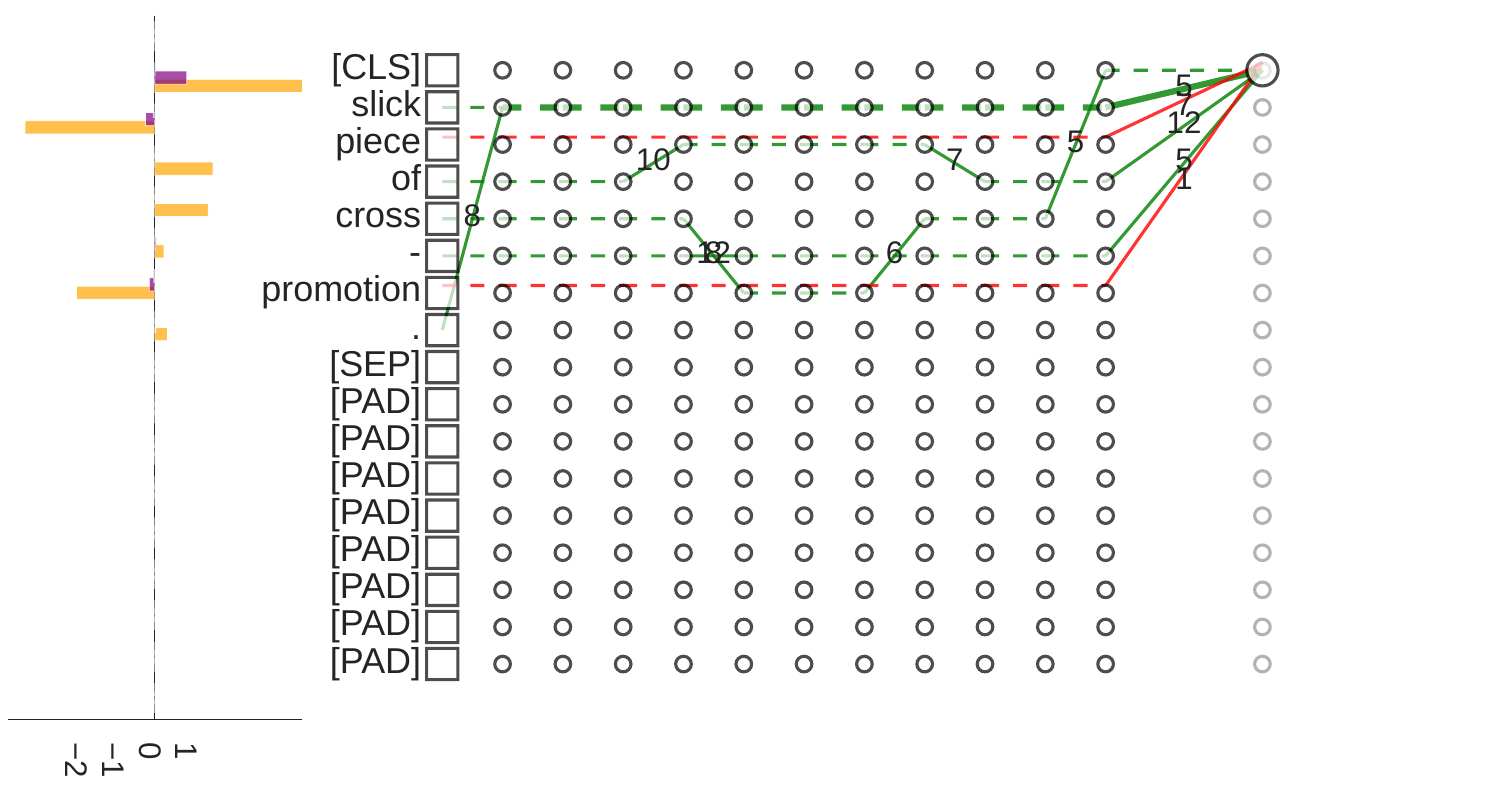}
	\caption{Example pattern of a positive sentence in SA}
	\label{fig:sst1}
\end{figure}

\begin{figure}[h]
	\centering
	\includegraphics[width=\linewidth]{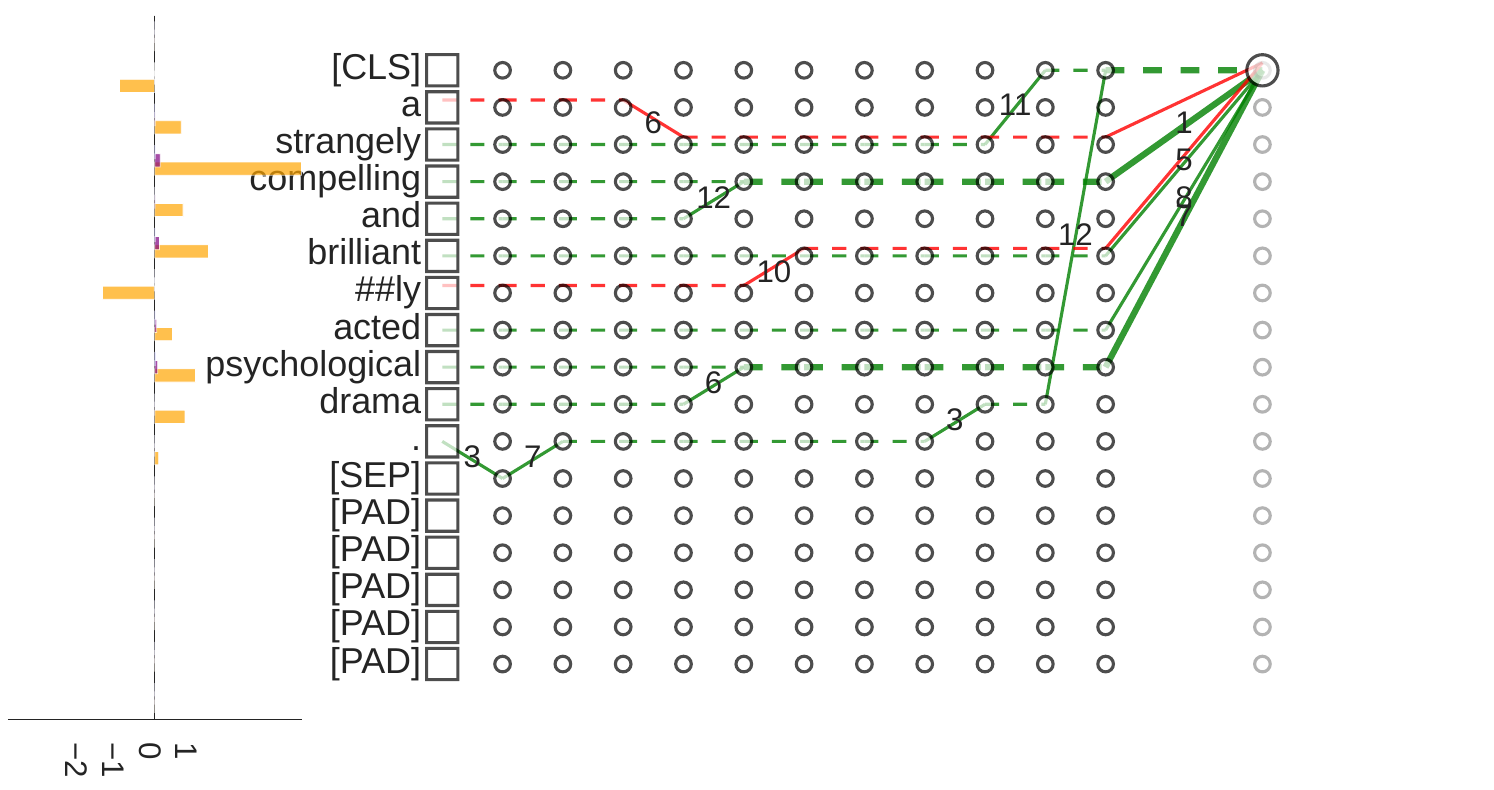}
	\caption{Example pattern of a positive sentence in SA}
	\label{fig:sst2}
\end{figure}

\begin{figure}[h]
	\centering
	\includegraphics[width=\linewidth]{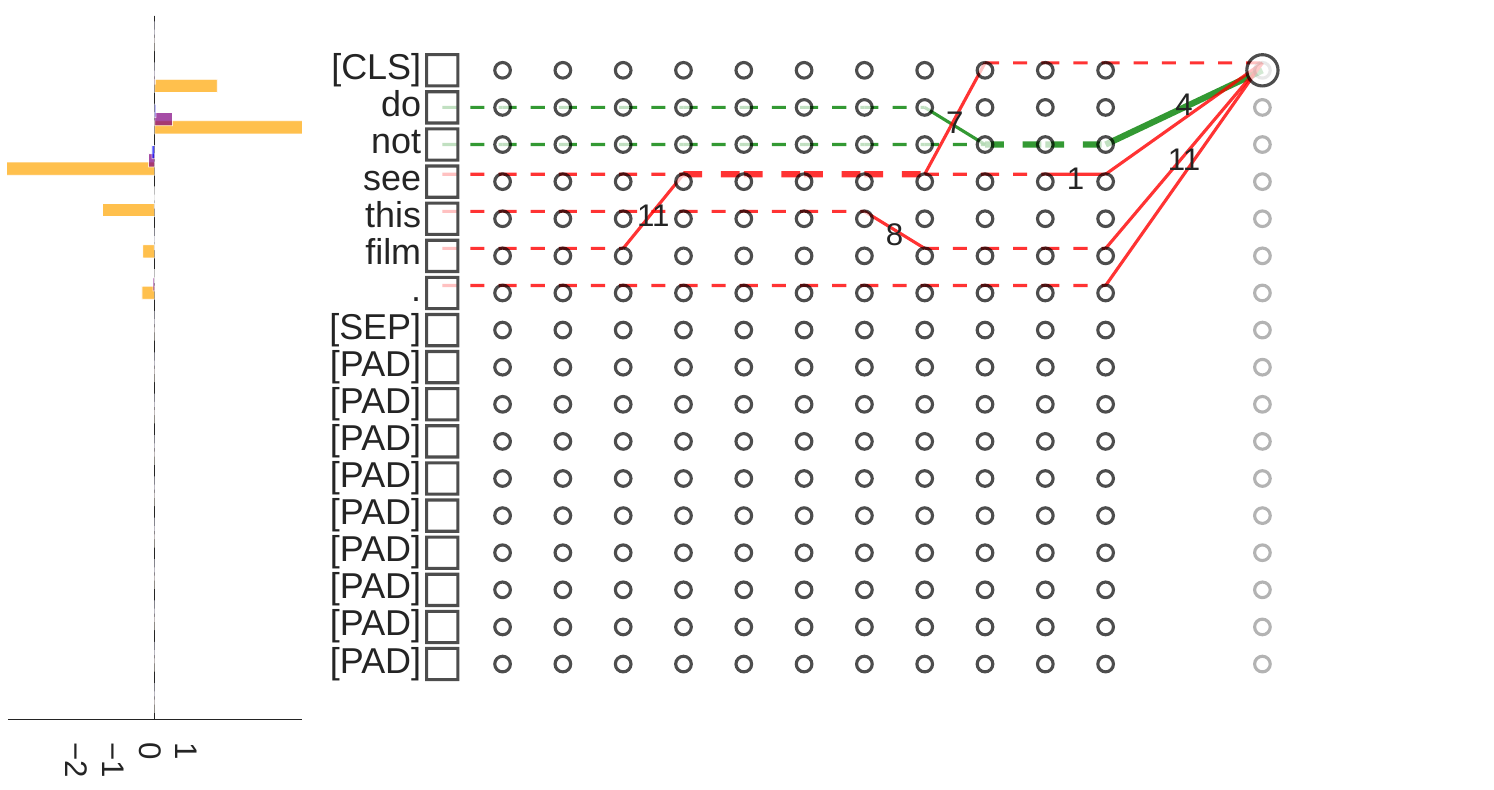}
	\caption{Example pattern of a negative sentence in SA}
	\label{fig:sst3}
\end{figure}

\begin{figure}[h]
	\centering
	\includegraphics[width=\linewidth]{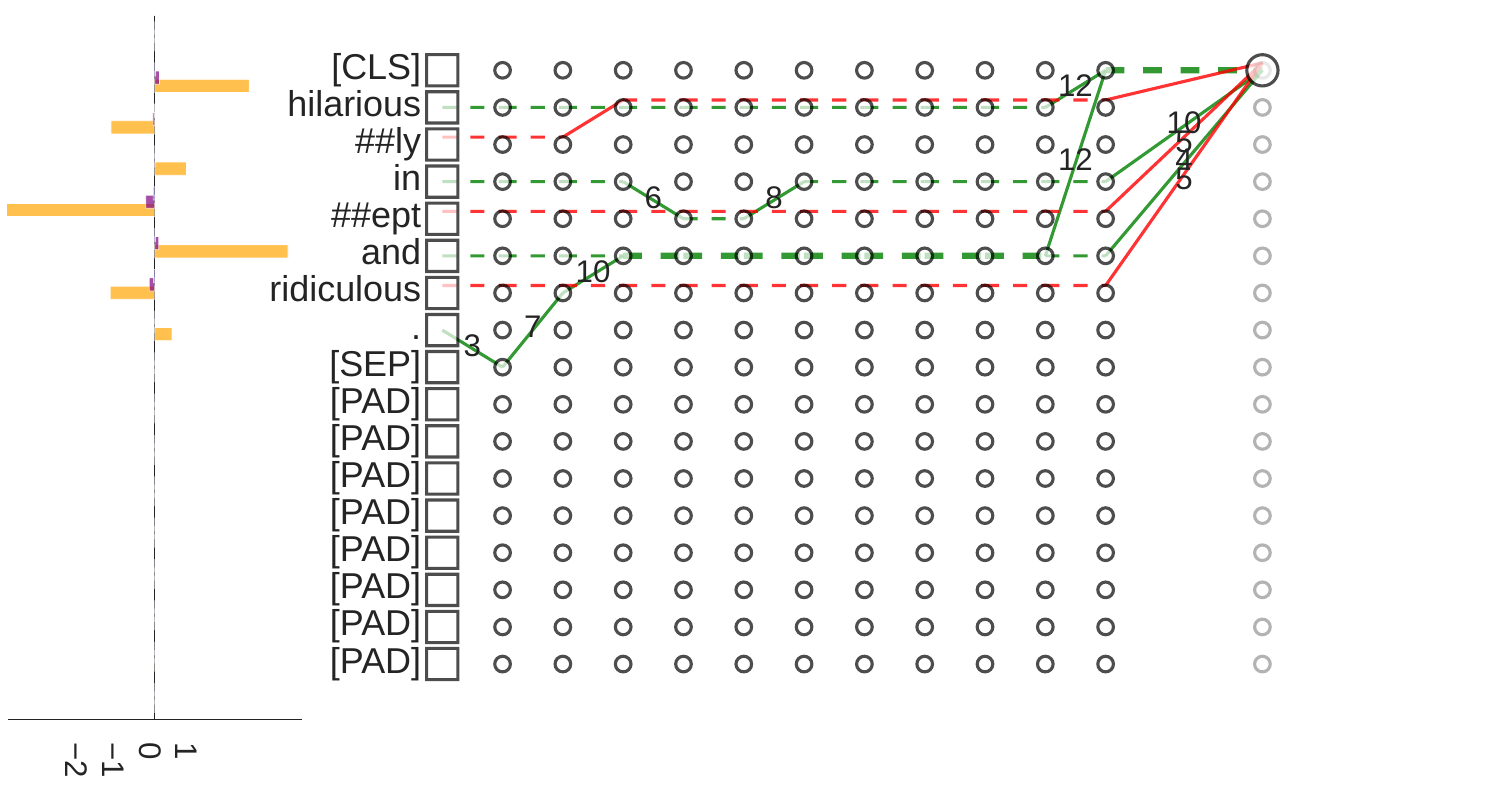}
	\caption{Example pattern of a positive sentence in SA}
	\label{fig:sst4}
\end{figure}

\begin{figure}[h]
	\centering
	\includegraphics[width=\linewidth]{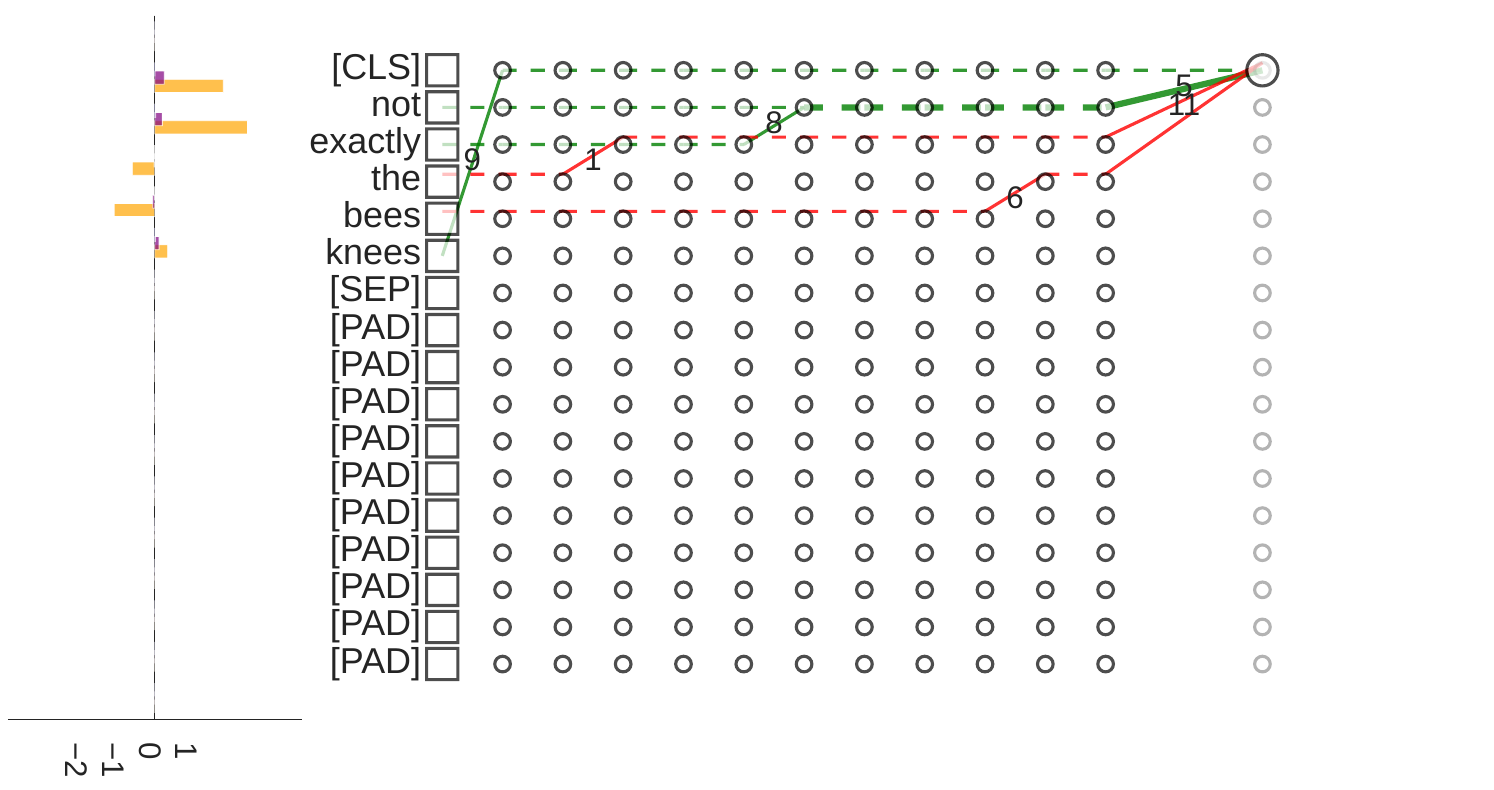}
	\caption{Example pattern of a negative sentence in SA}
	\label{fig:sst5}
\end{figure}

Figure~\ref{fig:e1},~\ref{fig:e2},~\ref{fig:e3} shows similar attractor examples from Figure~\ref{fig:gpr_ps} and ~\ref{fig:gpr_sp}, in three other evaluated subtask: SVA-Subj, SVA-APP, RA-NA. We observe similar discrepancies between SP and PS within each subtask, with \texttt{that}, and \texttt{across} and \texttt{that} functioning as attractors, respectively. Figure~\ref{fig:sst1} through~\ref{fig:sst5} show example patterns of actual sentences in the SST2 dataset. 

\subsection{Aggregated Visualizations}
\label{appendix: influence graphs}
\begin{figure}[h]

	\begin{subfigure}[t]{.4\textwidth}
		\centering
		\includegraphics[width=\linewidth]{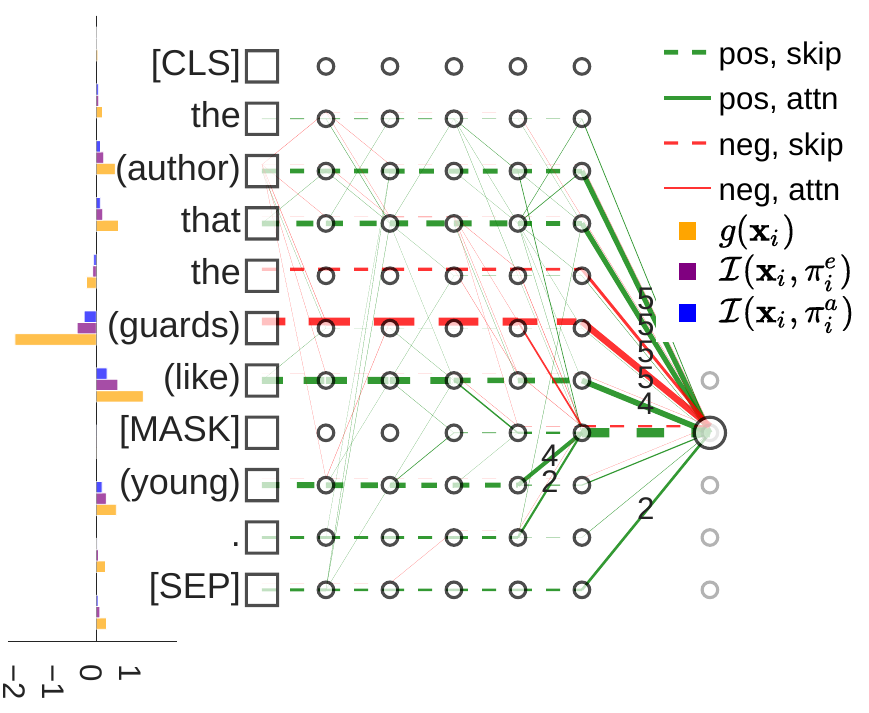}
		\caption{singular subject + plural intervening noun(SP)}
	\end{subfigure}
	\hfill
	\begin{subfigure}[t]{.4\textwidth}
		\centering
		\includegraphics[width=\linewidth]{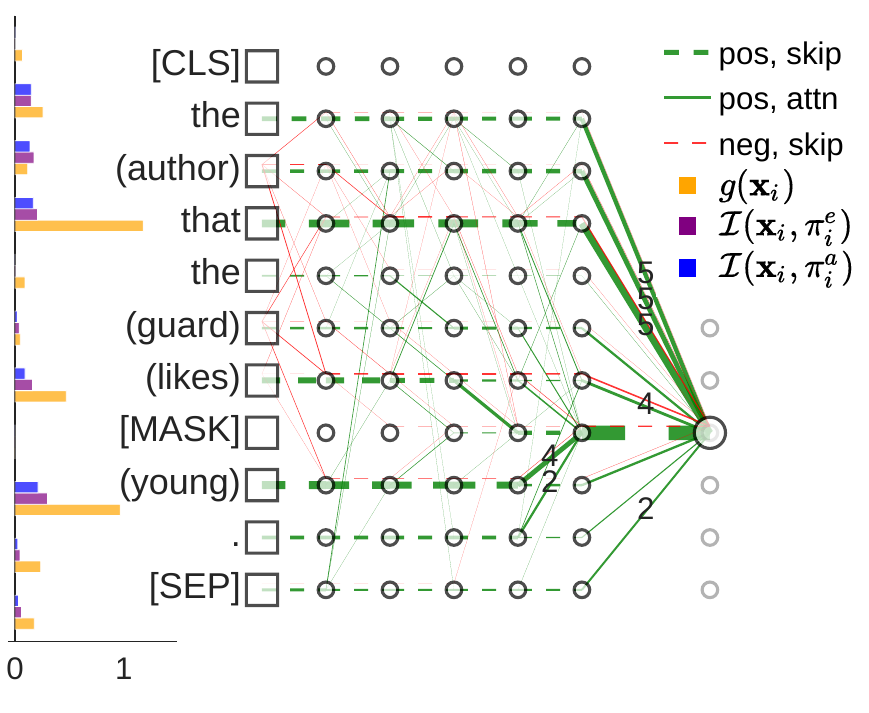}
		\caption{singular subject + singular intervening noun(SS)}
	\end{subfigure}

	\begin{subfigure}[t]{.4\textwidth}
		\centering
		\includegraphics[width=\linewidth]{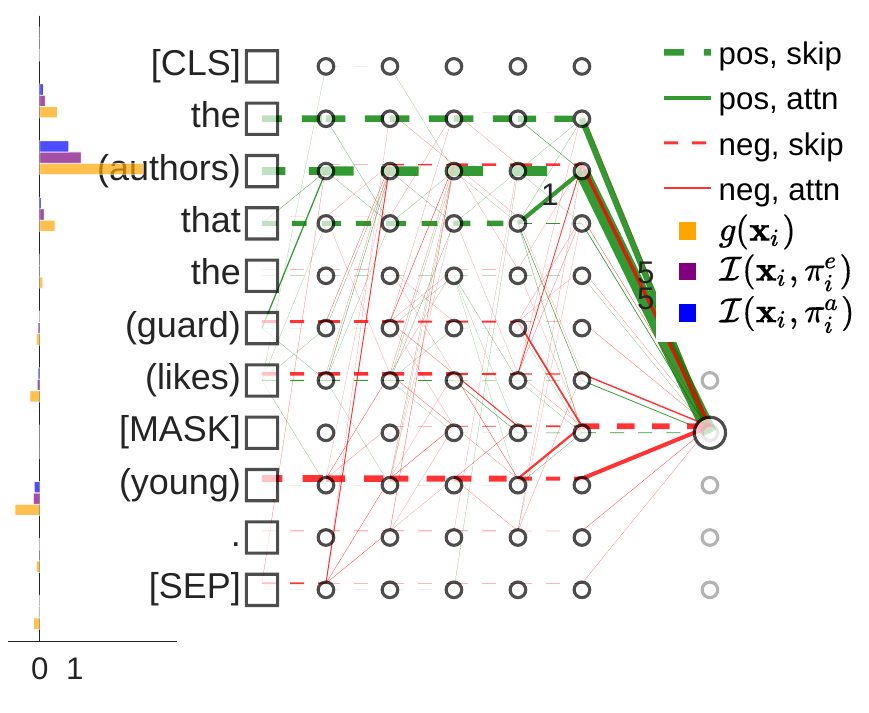}
		\caption{plural subject + singular intervening noun(PS)}
	\end{subfigure}
	\hfill
	\begin{subfigure}[t]{.4\textwidth}
		\centering
		\includegraphics[width=\linewidth]{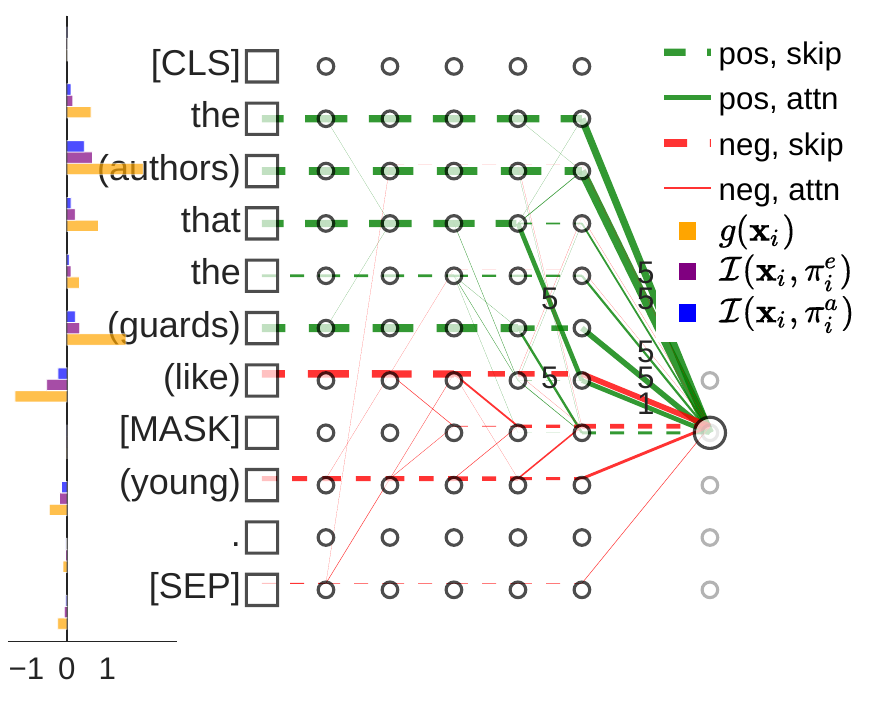}
		\caption{plural subject + plural intervening noun(PP)}
	\end{subfigure}

	%   \medskip

	\caption{SVA-Obj. Aggregated}
	\label{fig:e4}

\end{figure}

\begin{figure}[h]

	\begin{subfigure}[t]{.4\textwidth}
		\centering
		\includegraphics[width=\linewidth]{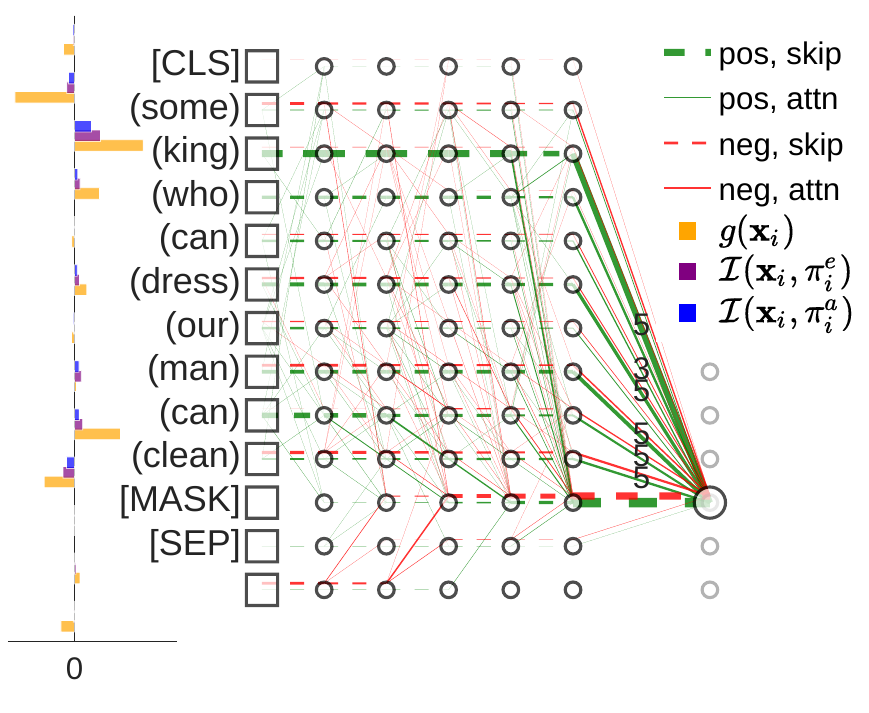}
		\caption{male subject + male intervening noun(MM)}
	\end{subfigure}
	\hfill
	\begin{subfigure}[t]{.4\textwidth}
		\centering
		\includegraphics[width=\linewidth]{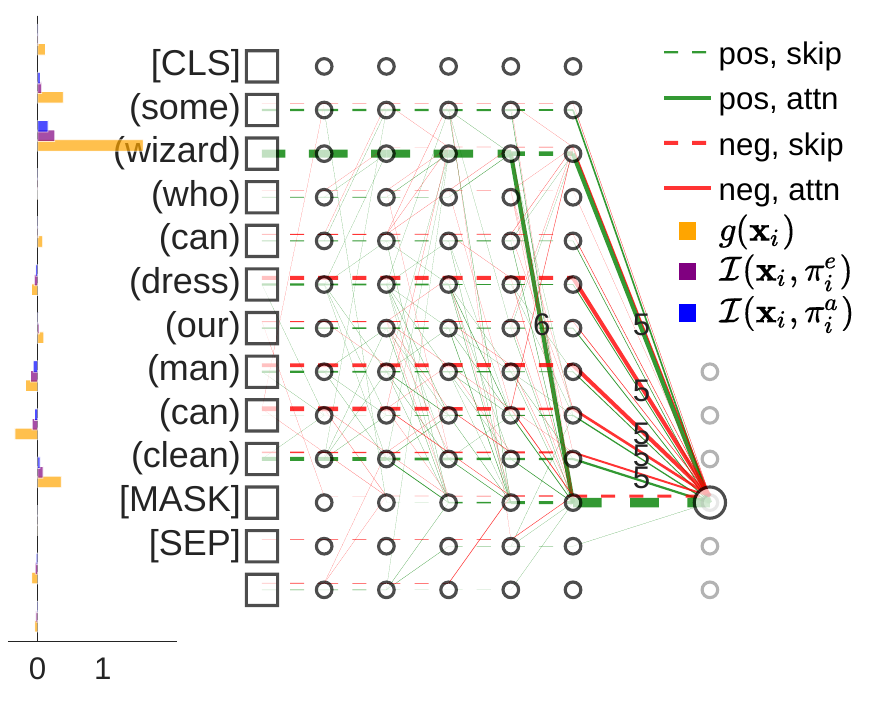}
		\caption{female subject + male intervening noun(FM)}
	\end{subfigure}

	\begin{subfigure}[t]{.4\textwidth}
		\centering
		\includegraphics[width=\linewidth]{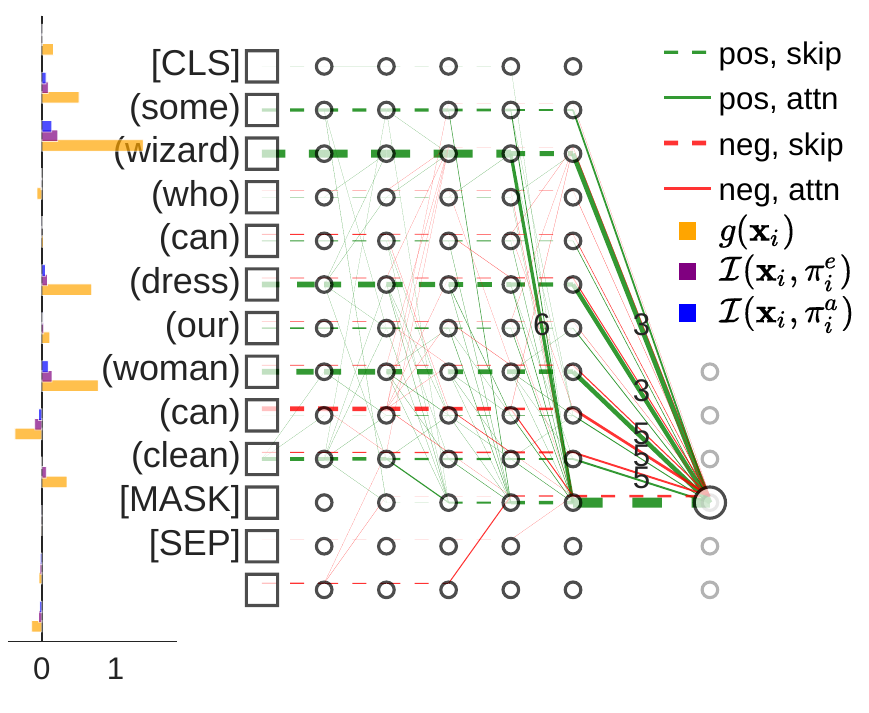}
		\caption{female subject + female intervening noun(FF)}
	\end{subfigure}
	\hfill
	\begin{subfigure}[t]{.4\textwidth}
		\centering
		\includegraphics[width=\linewidth]{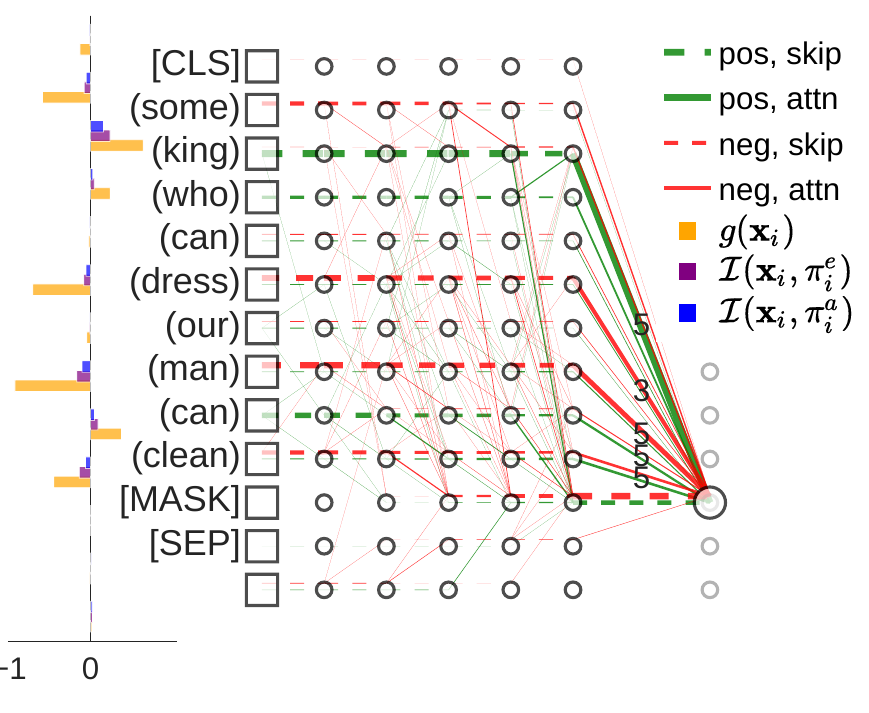}
		\caption{male subject + female intervening noun(MF)}
	\end{subfigure}
	\caption{RA: GA, Aggregated}
	\label{fig:e5}
\end{figure}

In this section, we show the aggregated visualization(Fig.~\ref{fig:e4}
and \ref{fig:e5}
) across all examples of each case in two subtasks (SVA-Obj \& NA-GA) by superimposing the patterns of individual instances (e.g. Figure~\ref{fig:gpr_ps} and ~\ref{fig:gpr_sp}), while adjusting the line width to be proportional to the frequency of flow across all examples. The words within parenthesis represent one instance of the word in that position. The aggregated graphs verify (1) generality of patterns across examples in each case (including SS and PP). (2) a more intuitive visualization of the pattern entropy in these two tasks, with RA-NA showing ``messier'' aggregated patterns, or larger entropy. 

\subsection{Impact Statement}
\label{appendix: impact}
Our work is expected to have general positive broader impacts on the uses of machine learning in the natural language processing. Specifically, we are addressing the continual lack of transparency in deep learning and the potential of intentional abuse of NLP systems employing deep learning. We hope that work such as ours will be used to build more trustworthy systems. Transparency/interpretability tools as we are building in this paper offer the ability for human users to peer inside the language models, e.g. BERT, to investigate the potential model quality issues, e.g. data bias and the abuse of privacy, which will in turn provide insights to improve the model quality. Conversely, the instrument we provide in this paper, when applied to specific realizations of language generation and understanding, can be used to scrutinize the ethics of these models' behavior. We believe the publication of the work is more directly useful in ways positive to the broader society. As our method does not use much resources for training or building new models for applications, we believe there are no significant negative social impacts.

%%%%%%%%%%%%%%%%%%%%%%%%%%%%%%%%%%%%%%%%%%%%%%%%%%%%%%%%%%%%

\end{document}